\documentclass[journal, twocolumn]{IEEEtran}

\makeatletter
\def\subsubsection{\@startsection{subsubsection}
                                 {3}
                                 {\z@}
                                 {0ex plus 0.1ex minus 0.1ex}
                                 {0ex}
                                 {\normalfont\normalsize\itshape}}
\makeatother

\ifCLASSINFOpdf
\else
\fi
\usepackage{times}
\usepackage{epsfig}
\usepackage{graphicx}
\usepackage{amsmath}
\usepackage{amssymb}

\usepackage{hyperref}       
\usepackage{url}            
\usepackage{booktabs}       
\usepackage{nicefrac}       
\usepackage{bm}
\usepackage{bbm}
\usepackage{amsmath}
\usepackage{subfig}
\usepackage{times}
\usepackage{epsfig}
\usepackage{multirow}
\usepackage{color}

\begin{document}

%
\title{Improved Techniques for Adversarial Discriminative Domain Adaptation}
%
%
%

\author{Aaron Chadha  and Yiannis Andreopoulos
\thanks{The authors are with the Electronic and Electrical Engineering Department, University College London, Roberts Building, Torrington Place, London, WC1E 7JE, UK (e-mail:
\{aaron.chadha.14, i.andreopoulos\}@ucl.ac.uk). The authors acknowledge support from the UK EPSRC, grants EP/R025290/1, EP/P02243X/1 and EP/R035342/1.}
\thanks{}
\thanks{}}

%
%

\markboth{to appear, ieee transactions on image processing}%
{Shell \MakeLowercase{\textit{et al.}}: Bare Demo of IEEEtran.cls for IEEE Journals}
%



\maketitle

\begin{abstract}
Adversarial discriminative domain adaptation (ADDA) is an efficient framework for unsupervised domain adaptation in image classification, where the source and target domains are assumed to have the same  classes, but no labels are available for the target domain. While ADDA has already achieved better training efficiency and competitive accuracy on image classification in comparison to other adversarial based methods, we investigate whether
we can improve its performance  with a new framework and new loss formulations.  Following the framework of  semi-supervised GANs, we first extend the discriminator output over  the source classes, in order to model the joint distribution over domain and task. We thus leverage on the distribution over the  source encoder posteriors (which is fixed during adversarial training) and propose maximum mean discrepancy (MMD) and reconstruction-based loss functions for aligning the target  encoder distribution to the source domain. We compare and provide a comprehensive analysis of how our framework and loss formulations extend over simple multi-class extensions of ADDA and other discriminative variants of semi-supervised GANs. In addition, we introduce various forms of regularization for stabilizing training, including treating the discriminator as a denoising autoencoder and regularizing the target encoder with source examples to reduce overfitting under a contraction mapping (i.e., when the target per-class distributions are contracting during alignment with the source). Finally, we validate our framework on standard datasets like MNIST, USPS, SVHN, MNIST-M and Office-31. We additionally examine how the proposed framework  benefits recognition problems based on sensing modalities that lack training data. This is realized by introducing and evaluating on a neuromorphic vision sensing (NVS) sign language recognition dataset,  where  the source domain constitutes emulated neuromorphic spike events converted from conventional pixel-based video and the target domain is experimental (real) spike events from an NVS camera. Our results on all datasets show that our proposal is both simple and efficient, as it competes or outperforms the state-of-the-art in unsupervised domain adaptation, such as DIFA and MCDDA, whilst offering lower  complexity than other recent adversarial methods. 


\end{abstract}


%
\IEEEpeerreviewmaketitle

\section{Introduction}

\IEEEPARstart{A} LONG-STANDING goal in visual learning is to generalize the learned knowledge from a source domain to new  domains, even without the presence of labels in the target domains. Significant strides have been made towards this goal in the last few years, mainly due to proposals based on multilayered convolutional neural networks that have shown cross-domain generalizations and fast learning of new tasks by fine-tuning on limited subsets of labelled data. 

Unsupervised domain adaptation directly aims at improving the generalization capability between  a labelled source domain and  an unlabelled target domain. Deep domain adaptation methods can generally be categorized as either being discrepancy based or adversarial based, with the common end goal of minimizing the difference between the  source and  target distributions.  Adversarial methods in particular have become increasingly popular due to their simplicity in training and success in minimizing the domain shift.  In this paper we focus on the recently proposed adversarial discriminative domain adaptation (ADDA) \cite{tzeng2017adversarial}, which is related to generative adversarial learning and uses the GAN \cite{goodfellow2014generative} objective to train on the target domain adversarially until it is aligned to the source domain.   ADDA uses the source image dataset labels solely for the pretraining of the source encoder. We improve upon this by: 
\begin{itemize}
\item extending the discriminator output over the source classes, in order to  additionally  incorporate task knowledge into the adversarial loss functions;
\item leveraging on the fixed distribution over source encoder posteriors in order to propose a maximum mean discrepancy (MMD) \cite{gretton2012kernel} and reconstruction-based loss function for training a target encoder  and discriminator, respectively;
\item comparing and providing a complete analysis of how our method extends over a simple multi-class extension of ADDA and discriminative variants of semi-supervised GANs \cite{odena2016semi, salimans2016improved}, including a    proposed pseudo-label  adversarial loss function that can be viewed as a multi-class version of the inverted label GAN\ setting \cite{goodfellow2014generative}.
\item addressing the issue of performance loss and overfittting for adversarial adaptation\ when the target class distributions are being contracted for source alignment; we refer to this case as a contraction mapping.
   
\end{itemize}

We benchmark the performance of our proposal against the state-of-the-art by evaluating on standard domain adaptation tasks with digits and Office-31 active pixel sensing (APS) datasets, showing that  we surpass the performance of ADDA by up to 21\% and remain competitive or superior to other recent proposals.
To highlight how our proposed framework can alleviate domain shift occurring in recognition tasks due to the use of novel sensing modalities, we select the emerging scenario of recognition tasks based on neuromorphic vision sensing (NVS)-based recognition. NVS hardware like the iniLabs DAVIS and the Pixium Vision\ ATIS\ cameras \cite{delbruck2010activity,neftci2015neuromorphic,benosman2014eventbasedvisualflow}  emulate the photoreceptor-bipolar-ganglion cell information flow and their output consists of asynchronous ON/OFF address events (a.k.a., spike events) that indicate the changes in scene reflectance. Existing NVS cameras can produce spike representations that can be rendered into frame representations comprising up to 2000 frames-per-second (fps), whilst operating with robustness to changes in lighting and at low power, on the order of 10mW. However, the events generated by NVS cameras are typically sparse and substantially more difficult to train on than APS\ domain inputs,  predominantly due to the lack of labelled NVS data currently available for training. We therefore introduce and evaluate on a new NVS  sign language recognition dataset, in which we present the emulated (source, labelled) $\rightarrow$ real (target, unlabelled) NVS domain adaptation task, showing substantial improvement on accuracy compared to ADDA and training on the source only.

The remainder of the paper is organized as follows.  Section \ref{sec:da_related_work} reviews recent work related to our proposals.
Section \ref{sec:improving_ADDA} introduces our proposed framework for improving ADDA, which constitutes our proposed loss formulations.  Section \ref{sec:semi} provides an extensive analysis of how we are able to bridge the gap from baseline ADDA with standard binary discriminator domain classification, to discriminative variants of semi supervised GANs and, finally, to our proposed loss formulations. In particular, we analyze our target encoder loss function and justify our design by comparing with conventional loss functions derived from semi-supervised GANs.  In Section \ref{sec:reg}, we consider various domain adaptation scenarios and introduce a method for target regularization with source examples. Finally, in Section \ref{sec:exp_results} we validate our proposal on conventional pixel domain datasets and our newly introduced emulated-to-real
NVS dataset for sign language recognition, and Section \ref{sec:conclusion} concludes the paper. 
\begin{figure*}
\centering
\includegraphics[scale=0.5]{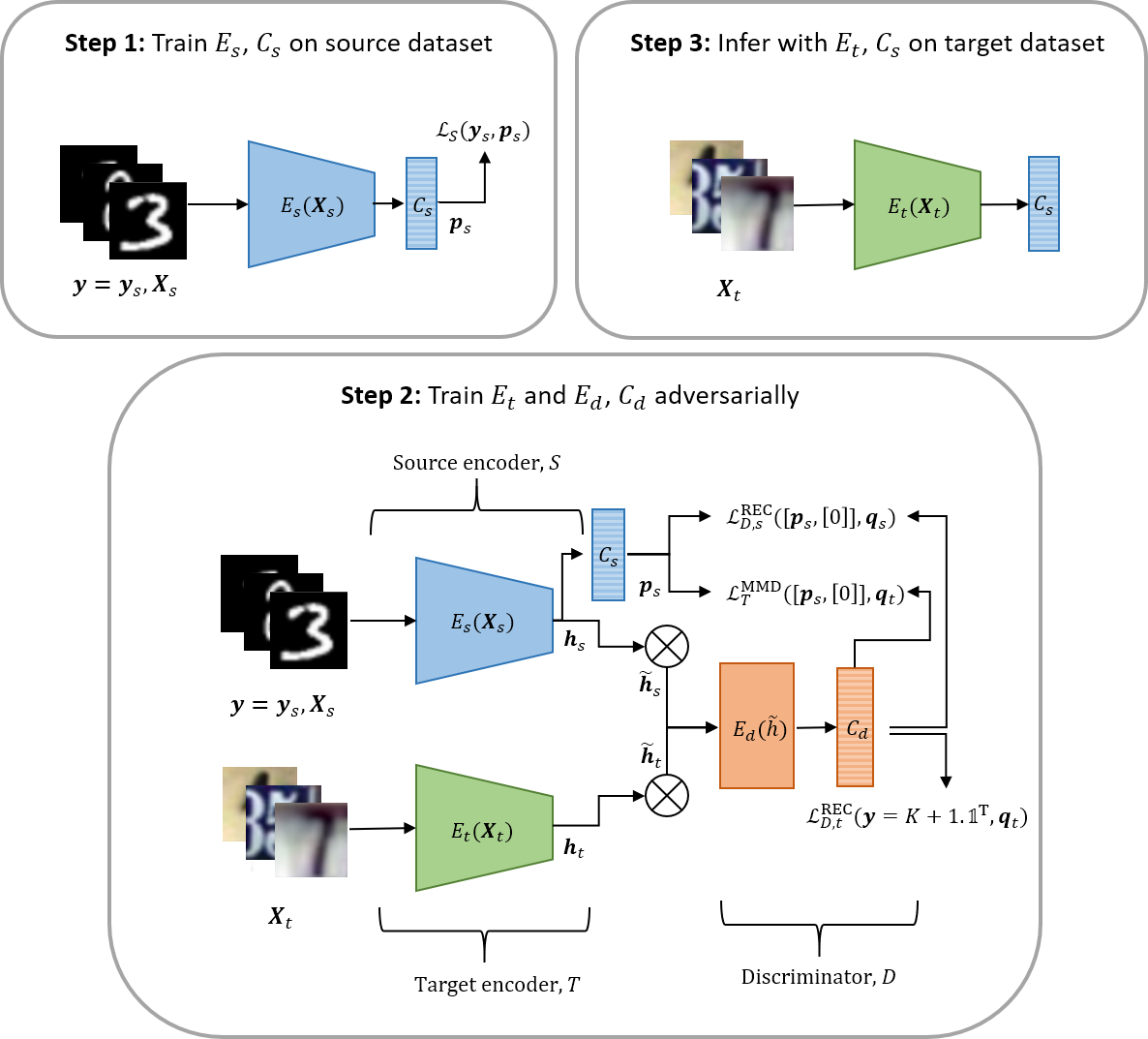}
\caption{Proposed improvements for adversarial discriminative domain adaptation. The figure shows the best configuration for training and inference explored in the paper.}
\label{fig:ss_adda}
\end{figure*}        
 
\section{Related Work}\label{sec:da_related_work}

We briefly discuss recent developments in deep learning for  unsupervised domain adaptation. In general, we can segment recent work into discrepancy based and adversarial based methods. We also review and illustrate the need for domain adaptation within the context of NVS based recognition systems.\\

\noindent\textbf{Discrepancy based methods:} Discrepancy  based methods  minimize the domain distribution discrepancy directly, typically  using an integral probability metric (IPM) based metric  such as   MMD \cite{gretton2012kernel} loss for this purpose.  MMD\ maps the original data   to  a  reproducing  kernel  Hilbert  space  (RKHS), where the source and target distributions are assumed separable.  Notably,  MMD is commonly used with a Gaussian kernel, which from the Taylor expansion enables  matching between all moments of distributions, albeit with some cost in processing. For example, Tzeng \textit{et al.} \cite{tzeng2014deep} proposed the deep domain confusion (DDC) method which applied  a joint classification and linear MMD loss on an intermediate adaptation layer. Long \textit{et al.} \cite{long2015learning} extended on DDC by adding multiple task-specific adaptation layers and minimizing the domain shift with a multiple-kernel maximum mean discrepancy.  Rather than matching the marginal distributions, the joint adaptation network (JAN) \cite{long2016deep} aligns the domain shift between the joint distributions of input features and output labels. More recently, Li \textit{et al.} \cite{li2018domain} proposed DICD, which uses MMD to match the marginal and conditional distributions in an iterative refinement manner.
     Alternatively, CORAL \cite{sun2016deep} matches only the mean and covariance between distributions, which despite its simplicity in only matching second order moments,  still maintains competitive performance. More recently,  Haeusser \textit{et al.} \cite{haeusser2017associative} proposed associative domain adaptation that replaces the MMD with an efficient discrepancy-based alternative that reinforces association between source and target embeddings. The basis of associativity is the two-step round-trip  probability  of  a  random  walker  starting
from a
labelled source feature and ending at another source feature via transition to a target feature. Associative cycle probabilities are encouraged to be close to a uniform distribution.     Effectively, associative domain adaptation uses the clustering assumption, where target and source samples from the same class should be located in high density regions of the feature space, with low density regions between classes. Similarly, this  assumption is adopted by Shu  \textit{et al.} \cite{shu2018dirt}, who add an additional penalty loss function to their adversarial learning framework, in order to punish violation of the clustering assumption. More recently, LDADA \cite{lu2018embarrassingly} uses the discrepancy between the per-class means for each domain to learn an LDA-like projection of the data. The target class assignment is estimated with pseudo-labelling. \\

\noindent\textbf{Adversarial based methods.}   In this paper, adversarial learning methods constitute the main point of comparison as our proposal directly improves on adversarial discriminative domain adaptation. Adversarial based methods opt for an adversarial loss function in order  to minimize the domain shift. The domain adversarial neural network (DANN) \cite{ganin2016domain} first introduced a gradient reversal layer that reversed the gradients of a binary classifier predicting the domain in order to train for domain confusion.  Training is performed jointly with a cross entropy loss that classifies the source examples, in order to learn a shared task-based embedding.  Other recent proposals \cite{liu2016coupled, bousmalis2017unsupervised, taigman2016unsupervised} have explored generative models such as  GANs \cite{goodfellow2014generative, mirza2014conditional} to learn from synthetic source and target data. These approaches typically train two GANs on the source and target input data with tied parameters.       In order to circumvent the need to generate images, Adversarial Discriminative Domain Adaptation (ADDA) \cite{tzeng2017adversarial} was recently proposed as an adversarial framework for directly minimizing the distance  between  the source and target encoded representations.  A discriminator and target encoder are iteratively optimized  in a two-player game akin to the original GAN\ setting, where the goal of the discriminator is to distinguish the target representation from the source domain and the goal of target encoder is to confuse the discriminator.  This implicitly aligns the target distribution to the (fixed) source distribution. The simplicity and power of ADDA\ has been demonstrated in visual adaptation tasks like MNIST, USPS\ and SVHN digits datasets.    The recently proposed DIFA \cite{volpi2018adversarial} extends this discriminative adversarial framework further by training a generator to generate source-like features that can be used to supplement the source examples during target adversarial training. Finally, Saito \textit{et al.}  \cite{saito2018maximum} propose  MCDDA, a feature generator and  two classifiers are trained in an adversarial fashion. This framework is based on the assumption  that target examples that fall outside the support of the source will be misclassified by two different classifiers. They alternately maximize a discrepancy based loss function to train the two classifiers and minimize the same function to train the generator, such that the generator will eventually generate target features that fall inside the support of the source and thus are more easily classified. Importantly, MCDDA has no source/target domain classification component and no explicit discriminator; on the contrary, our proposal embeds both a task classifier and domain classifier into the discriminator with a single head. We show that modelling the joint distribution over domain and task can improve performance. \\

\textbf{The need for domain adaptation in NVS domain.} A pertinent example of  a domain where it is difficult to obtain labelled data is neuromorphic vision sensing (NVS). NVS cameras produce coordinates and timestamps of on/off spikes in an asynchronous manner, i.e., when the logarithm of the intensity value of a CMOS sensor grid position changes beyond a threshold due to scene luminance changes. While such cameras are
now gaining traction as a low-power/high-speed visual sensing technology that circumvents
the limitations of conventional active pixel sensing (APS) cameras, there are currently very
limited or no annotations in the NVS domain for higher-level semantic tasks. This has been widely recognized \cite{delbruck2010activity} to be hampering progress in the adoption of NVS\ hardware within practical applications. Emulation from APS\ to NVS\ has attempted to provide for a solution \cite{bi2017pix2nvs}. 
Essentially, emulated NVS data is generated directly from APS frames, for which there is an abundance of publicly available datasets.  The APS instances are typically labelled and these can therefore be carried over to the emulated NVS domain, which now constitutes the source domain. Therefore, we can present the transductive transfer learning  from the labelled emulated NVS domain to the unlabelled real NVS\ domain  as an unsupervised domain adaptation problem. We evaluate our improved techniques for ADDA training on a new NVS sign language recognition task and demonstrate that our performance gains generalize to the NVS modality.

\section{Improving  Adversarial Adaptation} \label{sec:improving_ADDA}

\begin{table}
\noindent \centering\protect\caption{\label{tab:Notation-table.1}{Nomenclature Table.}}
\begin{tabular}[t]{cp{0.65\columnwidth}}
\toprule
\noalign{\vskip\doublerulesep}
\multicolumn{1}{c}{\textbf{Symbol }} & \multicolumn{1}{c}{\textbf{Definition}}
\tabularnewline[\doublerulesep]
\midrule
$\boldsymbol{x}_s, \boldsymbol{x}_t$  & Source / target images \tabularnewline
$K$ & Number of task-specific classes\tabularnewline
$y_s$  & Source task-specific  labels ($y_s \in \{1,\dots,K\}$)\tabularnewline
$y$  &  Extended class labels ($y \in \{1,\dots,K+1\}$)\tabularnewline
$E_s(.;\theta_s), E_t(.;\theta_t)$ & Source / target encoder function, parameterized by $\theta_s$ and $\theta_t$ respectively \tabularnewline
$E_d(.;\theta_d)$ & Discriminator encoder function, parameterized by $\theta_d$\tabularnewline
$C_s$ & Encoder classifier function (softmax) \tabularnewline
$C_d$ & Discriminator classifier function (softmax) \tabularnewline
$D$ & Complete discriminator mapping ($D= C_d(E_d)$) \tabularnewline  
$\boldsymbol{h}_s, \boldsymbol{h}_t$  & Source / target encoder logits ($\boldsymbol{h}=E(\boldsymbol{x})$) \tabularnewline
$\boldsymbol{\tilde{h}}_s, \boldsymbol{\tilde{h}}_t$  & Source / target corrupted encoder logits \tabularnewline
$N(\boldsymbol{\tilde{h}}|\boldsymbol{h})$ & Conditional distribution of corrupted encoder logits $\boldsymbol{\tilde{h}}$ given encoder logits  $\boldsymbol{h}$\tabularnewline
$z$  & Dropout keep probability for corruption process \tabularnewline
$\mathbf{X}_S,\mathbf{X}_T$ & Set of source image and label pairs / set of target images  \tabularnewline
$\mathbb{D}_S, \mathbb{D}_T$& Distribution over source image and label pairs / distribution over target images\tabularnewline
$\mathbb{H}_S, \mathbb{H}_T$ & Source / target distribution over encoder logit and label pairs and   logits respectively \tabularnewline
$\mathbb{P}_S, \mathbb{P}_T$ & Source / target distribution over encoder posteriors \tabularnewline
$\mathbb{Q}_S, \mathbb{Q}_T$ & Source / target distribution over discriminator   posteriors \tabularnewline

$\boldsymbol{p}_s , \boldsymbol{p}_t$ & Source / target encoder posteriors ($\boldsymbol{p} = C_s(\boldsymbol{h}$)) \tabularnewline 
$\boldsymbol{q}_s , \boldsymbol{q}_t$ & Source / target discriminator posteriors ($\boldsymbol{q}= D(\boldsymbol{\tilde{h}}$)) \tabularnewline
 
$\phi$ & Feature map to RKHS  ($\phi: \mathcal{X} \rightarrow \mathcal{H}$) \tabularnewline
$k$ & Radial basis function (RBF)  kernel  
\tabularnewline
$\sigma_r$ & Standard deviation of the $r$-th RBF kernel  
\tabularnewline
$\mathcal{F}$ & Function class for RKHS ($\mathcal{F}=\{f:\left\Vert f\right\Vert \leq 1\}$)
\tabularnewline
$\mathcal{H}$ & Reproducing kernel hilbert space (RKHS)  
\tabularnewline
$\boldsymbol{\hat{p}}_s,[\boldsymbol{p}_s,[0]]$ & Zero concatenated source encoder posteriors  \tabularnewline
$y_\mathrm{pred}$ & Predicted task label ($y_\mathrm{pred} \in \{1, \dots, K\}$) \tabularnewline
\bottomrule
\label{tab:nomenclature}
\end{tabular}
\end{table}

We illustrate  the framework for improving unsupervised adversarial discriminative domain adaptation in Fig. \ref{fig:ss_adda} and list all relevant symbols with their definitions in Table \ref{tab:nomenclature} for  reference purposes. Let $\mathbf{X}_S=\{(\boldsymbol{x}_s^i, y_s^i)\}^{N_s}_{i=0}$ represent the set of source image and label pairs, where $(\boldsymbol{x}_s, y_s) \sim \mathbb{D}_S$,   $\mathbf{X}_T=\{(\boldsymbol{x}_t^i)\}^{N_t}_{i=0}$ represent the set of unlabeled target images, $\boldsymbol{x}_t \sim \mathbb{D}_T$.  Let $E_s(\boldsymbol{x}_s;\theta_s)$ represent the source encoder function, parameterized     by $\theta_s$ which maps an image $\boldsymbol{x}_s$ to the encoder  output $\boldsymbol{h}_s$, where $(\boldsymbol{h}_s,y_s)\sim \mathbb{H}_{S}$. Likewise, let $E_t(\boldsymbol{x}_t;\theta_t)$ represent the target encoder function, parameterized     by $\theta_t$ which maps an image $\boldsymbol{x}_t$ to the encoder  output $\boldsymbol{h}_t$, where $\boldsymbol{h}_t \sim \mathbb{H}_T$.  In addition,  $C_s$ represents a classifier function that maps the encoder output $\boldsymbol{h}$ to class probabilities $\boldsymbol{p}$.  In this paper, we   only consider   $\boldsymbol{h}_s$ and $\boldsymbol{h}_t$ as representing  the source and target  logits respectively  and therefore $C_s$ simply denotes the softmax function on the logits. Finally, let $E_d(\boldsymbol{h};\phi_d)$ represent an encoder mapping from $\boldsymbol{h}$ to an intermediate representation, and $C_d$ represent a softmax function on said representation; $E_d$ and $C_d$ jointly constitute our discriminator mapping, which we refer to as  $D = C_d(E_d)$.

 Our  objective is to substantially improve the  adversarial training of the target encoder in ADDA \cite{tzeng2017adversarial}. Rather than training the discriminator $D$ and target encoder  $E_t$ with the standard GAN\ loss formulations (i.e., training a logistic function on the discriminator by assigning labels 1 and 0 to the source and  domains respectively and training the generator with inverted labels \cite{goodfellow2014generative}), our approach is inspired by semi-supervised  GANs \cite{odena2016semi, salimans2016improved}, where it has been found that incorporating task knowledge into the discriminator can jointly improve classification performance and quality of images produced by the generator. Under the discriminative adversarial framework, we can equivalently incorporate task knowledge by replacing the discriminator  logistic function with a $K+1$ multi-class classifier.
However, unlike  the GAN setting, the discriminator inputs and outputs can now both be represented with  $K+1$ dimensions, with each dimension representing a class; we leverage on this fact in our proposed\ loss formulations in Sections \ref{sec:discriminator} and \ref{sec:target_enc} to improve the convergence properties of our framework, in comparison to the original ADDA proposal \cite{tzeng2017adversarial}.

 We begin by outlining   three main steps for our proposed adversarial framework, which involve learning the source mapping on the source dataset, adversarial training to align the source and  target domains and finally inferring on the target dataset.  The classifier $C_s$ is fully interchangeable between the source encoder $E_s$ and the target encoder $E_t$.  This means we can embed $C_s$ into the adversarial training of the target encoder $E_t$ and discriminator $D$.

\subsection{Step 1: Supervised Training of the Source Encoder and Classifier}  Given that we have access to labels in the source domain, we first train the source encoder $E_s$ and classifier $C_s$ on the source image and label pairs ($\boldsymbol{x}_s$, $y_s \in \{1,...,K\}$) in a supervised fashion, by minimizing the standard cross entropy loss with $K$ classes:

\begin{equation}
{L}_S = - \mathbb{E}_{{(\boldsymbol{x}_s, y_s)} \sim \mathbb{D}_S}\sum^{K}_{k=1} 1_{[k=y_s]}\log{(C_s(E_s(\boldsymbol{x}_s))_{k})}
\end{equation}

 The source encoder parameters $\theta_s$ are now frozen, which fixes the  distribution $\mathbb{H}_S$.  This becomes our reference distribution for adversarial training, analogous to the real image distribution in the GAN setting, where our aim is now to align the target distribution $\mathbb{H}_T$ to $\mathbb{H}_S$ by learning a suitable target encoding $E_t$.                                          

\subsection{Step 2: Adversarial Training of the Encoder}
\label{sec:step_2}

\noindent
\subsubsection{Discriminator loss function $L^\mathrm{REC}_D$} \label{sec:discriminator}

We train a target encoder adversarially  by passing the source and target encoder logits, $\boldsymbol{h}_s$ and   $\boldsymbol{h}_t$, to a discriminator $D$.    In doing so, we implicitly align the target encoder  distribution to that of the source; i.e., $E_t(\boldsymbol{x}_t) \sim \mathbb{H}_S$.  As the source encoder has fixed parameters, we  learn an asymmetric  encoding with untied weights, which is the standard setting in both ADDA \cite{tzeng2017adversarial} and GAN\ implementations \cite{goodfellow2014generative, mirza2014conditional}. In addition, we can improve the convergence properties by first  initializing  the target encoder weights with the source encoder  weights; i.e.,  $\theta_{t}$ = $\theta_{s}$.
This choice of initialization is motivated by GANs. As theoretically proven by Arjovsky and Bottou \cite{arjovsky2017towards} for the GAN setting, if the `real' and `fake' distributions are disjoint, we are always capable of finding an optimal discriminator and this leads to instability or vanishing gradients propagated to the generator. By initializing the target encoder weights with the source encoder weights, we ensure the target encoder distribution is not initially disjoint from the source  encoder distribution and that there is some initial clustering based on class, which thus stabilizes convergence.  

We extend the discriminator output  $\boldsymbol{q}  $ to a $K+1$ dimensional vector representing the class probabilities, in which the first  $K$ dimensions represent the joint distribution over   source domain    and the task specific classes and  final $K+1$-th dimension represents the target domain. We denote the $K+1$ class labels as  $y \in \{1,\dots,K+1\}$, where each source encoder logit $\boldsymbol{h}_s$ is assigned its task label $y = y_s   \in \{1,\dots,K\}$ and the `target domain' label $y=K+1$ is only assigned to target encoder logits  $\boldsymbol{h}_t$.  However, contrary to semi-supervised GANs where the discriminator inputs are images, the discriminator inputs and outputs now share common    supports over the $K$ task classes. For the source domain, we can leverage on this fact by effectively modelling the discriminator as a denoising autoencoder \cite{vincent2008extracting}, where we can jointly train the discriminator to reconstruct the source encoder logits and  encourage the discriminator  to potentially learn a more informative representation by corrupting the logits.  A denoising autoencoder is an effective method of approximating the underlying source logit  manifold and ensures that the discriminator  deviates away from learning a simple identity function. We refer to the corruption process as $N(\boldsymbol{\tilde{h}}_s|\boldsymbol{h}_s)$, which represents the conditional distribution over the corrupted source encoder logits $\boldsymbol{\tilde{h}}_s$ given  the source encoder logits $\boldsymbol{h}_s$. Therefore, the first term of our discriminator loss function is effectively a reconstruction loss, which we set as the  cross entropy between the  zero-concatentated source encoder posteriors $\boldsymbol{\hat{p}}_s = [C_s(\boldsymbol{h}_s),[0]]= [\boldsymbol{p}_s,[0]]$ and source discriminator posteriors $\boldsymbol{q}_s = C_{d}(E_{d}(\boldsymbol{\tilde{h}}_s;\phi_d))=D(\boldsymbol{\tilde{h}}_s)$ (i.e., post-softmax):

\begin{equation}
\begin{aligned}
 {L}^\mathrm{REC}_{{D},{s}} & = -\mathbb{E}_{{(\boldsymbol{h}_s, y_s) \sim\mathbb{H}_S}} \mathbb{E}_{\boldsymbol{\tilde{h}}_s \sim N(\boldsymbol{\tilde{h}}_s|\boldsymbol{h}_s)} ([C_s(\boldsymbol{h}_s),[0]] \cdot \log(D(\boldsymbol{\tilde{h}}_s))) \\& =-\mathbb{E}_{{(\boldsymbol{h}_s, y_s) \sim\mathbb{H}_S}} \mathbb{E}_{\boldsymbol{\tilde{h}}_s \sim N(\boldsymbol{\tilde{h}}_s|\boldsymbol{h}_s)}\sum^{K}_{k=1}\hat{p}_{s,k} \log (q_{s,k})  
\label{D1}
\end{aligned}
\end{equation}

\noindent where $p_{s,k}$ and $q_{s,k}$ are the $k$-th elements of $\boldsymbol{\hat{p}}_s$ and $\boldsymbol{q}_s$ respectively. Notably, we append a zero to the source encoder posteriors to represent  the $K+1$-th  `target domain' class, which maintains a valid probability distribution (sums to 1), whilst enforcing a zero probability that the posteriors were generated by the target encoder.  In this paper, the corruption process $N$ is simply configured as dropout on the encoder logits.

We also apply dropout independently  to the target encoder logits  $\boldsymbol{h}_t$, in order to symmetrize the source and target encoder inputs presented to the discriminator.  However, we want  the discriminator to distinguish between the source and target encoder logits.  We  train the discriminator to assign the $K+1$-th `target domain' class  to the corrupted target encoder logits $\boldsymbol{\tilde{h}}_t$, such that they lie in an orthogonal space to the source domain.   In other words, the second term of our  discriminator loss function for the target encoder logits is:

\begin{equation}
{L}^\mathrm{REC}_{D,t}  =  -\mathbb{E}_{{\boldsymbol{h}_t \sim\mathbb{H}_T}} \mathbb{E}_{\boldsymbol{\tilde{h}}_t \sim N(\boldsymbol{\tilde{h}}_t|\boldsymbol{h}_t)} \log(D(\boldsymbol{\tilde{h}}_t)_{K+1})
\label{D2}
\end{equation}

\noindent where $D(\boldsymbol{\tilde{h}}_t)_{K+1}$ is the $K+1$-th dimension of $D(\boldsymbol{\tilde{h}}_t)$. The discriminator loss function ${L}^\mathrm{REC}_D$ is thus simply the sum of  (\ref{D1}) and (\ref{D2}): ${L}^\mathrm{REC}_D={L}^\mathrm{REC}_{D,s} + {L}^\mathrm{REC}_{D,t}$.
In order to further motivate this reconstruction based loss function, we  derive a loss function akin to a discriminative variant to semi-supervised GANs in Section \ref{sec:onevstwo_heads} and compare  with our proposed formulation.

\ \subsubsection{Target encoder loss function ${L}^\mathrm{MMD}_T$} 
\label{sec:target_enc}

   In order to train the target encoder adversarially, we want the target encoder to generate an output that is representative of one of the first $K$ task-specific classes rather than the $K+1$-th `target domain' class that it is assigned when training the discriminator.  To achieve this, we leverage on the two source posteriors,  $\boldsymbol{p}_s=C_s(\boldsymbol{h}_s)$ and $\boldsymbol{q}_s=D(\boldsymbol{\tilde{h}}_s)$, generated by the source encoder and discriminator respectively.  Contrary to supervised domain adaptation methods,  there are no known source and target pairwise correspondences and we cannot formulate a paired test over the posteriors. However, we  can formulate the problem as a two-sample test by considering   the distribution over   target discriminator posteriors, $\boldsymbol{q}_t=D(\boldsymbol{\tilde{h}}_t)$, compared to the distribution over the source encoder posteriors  $\boldsymbol{p}_s$, where our null hypothesis is that the distributions are equal.   We consider a set of target discriminator posteriors $\mathbf{Q}_T = \{\boldsymbol{q}^1_t,\dots,\boldsymbol{q}^{n}_t\} \sim \mathbb{Q}_T$ and a set of source encoder posteriors $\mathbf{P}_S = \{\boldsymbol{p}_s^1,\dots,\boldsymbol{p}_s^{n}\} \sim \mathbb{P}_S$, where $n$ is the set cardinality  and $\mathbb{P}_S$ and $\mathbb{Q}_T$ are the respective posterior distributions.  Effectively, we want to minimize the distance between $\mathbb{P}_S$ and $\mathbb{Q}_T$ without performing any density estimation. To this end, we adopt the Maximum Mean Discrepancy (MMD) \cite{gretton2012kernel} metric as a measure of distance between the mean embeddings of $\boldsymbol{p}_s$ and $\boldsymbol{q}_t$. For  reproducing kernel Hilbert space (RKHS) $\mathcal{H}$, function class $\mathcal{F}$ = $\{f:\left\Vert f\right\Vert \leq 1\}$ and infinite dimensional feature  map $\phi : \mathcal{X} \rightarrow\ \mathcal{H}$ the MMD can be expressed as:                             

\begin{equation}
\begin{aligned}
\mathcal{D}_\mathrm{MMD} & = \sup_{f \in \mathcal{F}, \left\Vert f \right\Vert_{\mathcal{H} \leq 1}}\left\vert\mathbb{E}_{\boldsymbol{p}_s \sim \mathbb{P}_S }f([\boldsymbol{p}_s,[0]])- \mathbb{E}_{\boldsymbol{q}_t \sim \mathbb{Q}_T }f(\boldsymbol{q}_t)\right\vert \\ &= \left\Vert \mathbb{E}_{\boldsymbol{p}_s \sim \mathbb{P}_S}\phi([\boldsymbol{p}_s,[0]]) - \mathbb{E}_{\boldsymbol{q}_t \sim \mathbb{Q}_T}\phi(\boldsymbol{q}_t) \right\Vert_\mathcal{H}
\label{MMD}
\end{aligned}
\end{equation} 

The distribution $\mathbb{P}_S$ over source encoder posteriors  is fixed during adversarial training and, as such, we are effectively aligning the distribution $\mathbb{Q}_T$ over target discriminator posteriors to $\mathbb{P}_S$.    We again append a 0 to the source encoder posteriors to represent the `target domain' class probability, such that both source and target  posteriors are $K+1$ dimensional prior to mapping to $\mathcal{H}$. This zero constraint on the $K+1$-th class acts as a stronger prior upon which to learn the target encoder; as such,  the  source encoder posterior provides a more informative representation  than the source discriminator posterior.   The feature map $\phi$ in  (\ref{MMD}) corresponds to a positive semi-definite kernel $k$ such that $k(\boldsymbol{x},\boldsymbol{y}) = \left\langle \phi(\boldsymbol{x}),\phi(\boldsymbol{y}) \right\rangle_\mathcal{H}$, which means we can rewrite  (\ref{MMD}) in terms of $k$.  The loss function on our target encoder that we wish to minimize can thus be written for aligning $\mathbb{Q}_T$ to $\mathbb{P}_S$ as:      

\begin{equation}
\begin{aligned}
{L}^{\mathrm{MMD}}_{T}(\mathbb{P}_S \rightarrow \mathbb{Q}_T) & = {{D}_\mathrm{MMD}}^2 \\ & =   \mathbb{E}_{\boldsymbol{p}_s, \boldsymbol{p}_s' \sim \mathbb{P}_S, \mathbb{P}_S}k([\boldsymbol{p}_s,[0]],[\boldsymbol{p}_s',[0]]) \\ &- \mathbb{E}_{\boldsymbol{p_{s}}, \boldsymbol{q}_t \sim \mathbb{P}_S,  \mathbb{Q}_T}k([\boldsymbol{p}_s,[0]],\boldsymbol{q}_t) \\ & + \mathbb{E}_{\boldsymbol{q}_t, \boldsymbol{q}_t' \sim \mathbb{Q}_T, \mathbb{Q}_T}k(\boldsymbol{q}_t, \boldsymbol{q}'_t)
\label{target_loss_2}
\end{aligned}
\end{equation}

 In this paper we opt to use a linear combination of $r$ multiple radial basis function (RBF) kernels over a range of standard deviations, such that $k(\boldsymbol{x},\boldsymbol{y})=\sum_r \exp\{-\frac{1}{2\sigma_r}\left\Vert \boldsymbol{x}-\boldsymbol{y} \right\Vert^2_2\}$, where $\sigma_r$ is the standard deviation of the $r$-th RBF kernel.  We find that the standard  RBF kernel with squared Euclidean distance as above performs better in practice than a generalized RBF kernel with a  distribution based  metric such as chi-squared distance or squared Hellinger's distance - we present some representative  results on standard datasets in Appendix \ref{App A} to validate our choice of kernel function. By introducing a linear combination over varying  bandwidths, we improve the generalization performance over different sample distributions.  This method of generalization with fixed kernels  is commonly used both in generative models \cite{li2015generative, dziugaite2015training} and other  domain adaptation discrepancy based methods \cite{long2015learning, bousmalis2016domain}.    In order to further motivate our proposed MMD\ loss formulation, we introduce alternative target encoder loss functions in Section \ref{sec:semi} and a full ablation study on all introduced  discriminator-encoder loss combinations in Section \ref{sec:ablation}.

\subsection{Step 3: Inference on the Target Dataset}                     
 After training the target encoder, we can now perform inference on the target dataset. However, we have effectively trained two sets of target predictions; namely, the mapped  target encoder output $C_s(\boldsymbol{h}_t)$ and the  discriminator  output $\boldsymbol{q}_t$.  In the optimal setting, where we have trained the discriminator to equilibrium, we would expect  the discriminator mapped source and target distributions would be aligned.      However, we empirically find that evaluation  on $\boldsymbol{q}_t$  is marginally worse (\textasciitilde1\%)\ than evaluation on $C_s(\boldsymbol{h}_t)$.  Therefore, for the remainder of the paper, we infer on the target encoder output. The class prediction  $y_\mathrm{pred}$ is given as: 

\begin{equation}
y_\mathrm{pred} = \arg \max_{j \in \{1,\dots,K\}} (h_{t,j} )  
\end{equation}

\section{Bridging the Gap from ADDA to Our Proposal} \label{sec:semi}

We provide further insight on the design of our adversarial loss formulations by first demonstrating in Section \ref{sec:onevstwo_heads} how we can extend from ADDA \cite{tzeng2017adversarial}, to a multi-class version of  ADDA with separate task and domain  classification heads  and, finally, to a framework  with a single classification head. For the latter,   we perform a detailed comparison between the target encoder loss function in our proposal and  discriminative variants of semi-supervised GANs \cite{odena2016semi, salimans2016improved} in Section \ref{sec:semivsprop}.

\begin{figure*}
\centering

\subfloat[$\mathrm{ADDA}$]{\includegraphics[scale=0.7]{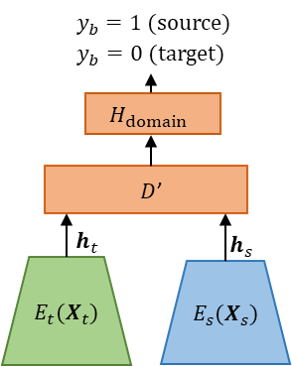}} \hspace{10pt} \subfloat[$\mathrm{MULTI}$]{\includegraphics[scale=0.7]{{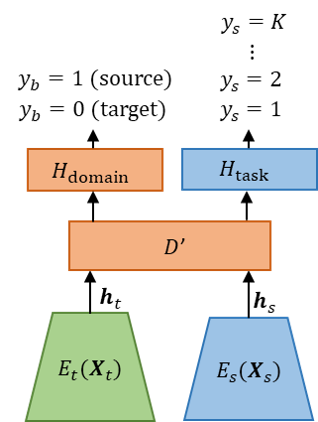}}} \hspace{10pt} \subfloat[$\mathrm{JOINT}$]{\includegraphics[scale=0.7]{{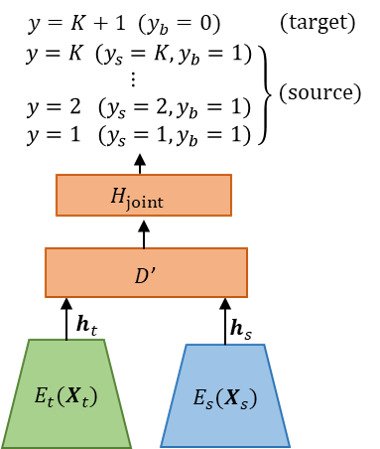}}}
\caption{Variants of discriminator configuration for adversarial training. (a)$\mathrm{ADDA}$: this represents the standard ADDA formulation with a single discriminator head for classifying the domain - this is the discriminative variant of the vanilla GAN. (b)$\mathrm{MULTI}$: this extends (a) with an additional task classification head $H_\mathrm{task}$ for classifying the source examples - this is effectively equivalent  to DANN \cite{ganin2016domain} (replacing the gradient reversal layer with an adversarial loss)  and represents a discriminative variant of the AC-GAN \cite{odena2017conditional}. (c) $\mathrm{JOINT}$: our proposal is adapted from this formulation, in which we learn a joint distribution over task and domain with a single discriminator head $H_\mathrm{joint}$ - this represents a discriminative variant of semi-supervised GANs \cite{odena2016semi, salimans2016improved}.         }
\label{fig:D_config}
\end{figure*}
  
\subsection{Transitioning from Two Heads to One Head} \label{sec:onevstwo_heads}
Let us denote a discriminator  classification head as the  layer $H$ and the preceding discriminator layers as $D'$. We begin with ADDA, which is typically trained with  a  domain classification head $H_\mathrm{domain}$, in which    the discriminator assigns a domain label  $y_b \in \{0,1\}$ to instances (where 1 corresponds to the source domain instance and 0 to the target domain instance). This configuration, which is effectively a discriminative variant of the vanilla GAN,   is illustrated in Fig. \ref{fig:D_config}(a). The discriminator loss function can be written as \cite{tzeng2017adversarial}:

\begin{equation}
\begin{aligned}
L^\mathrm{ADDA}_{D',H_\mathrm{domain}}=  & - \mathbb{E}_{(\boldsymbol{h}_s, y_s) \sim \mathbb{H}_S}\log(p_{\mathrm{domain}}(y_{b}=1|\boldsymbol{h}_s)) \\ &  - \mathbb{E}_{(\boldsymbol{h}_t) \sim \mathbb{H}_T}\log(1-p_{\mathrm{domain}}(y_{b}=1|\boldsymbol{h}_t))
  \label{eq_ADDA}
\end{aligned}  
\end{equation}

\noindent where $p_{{\mathrm{domain}}}(y_b=1|\boldsymbol{h})$      is the posterior probability output by $H_\mathrm{domain}(D'(\boldsymbol{h}))$   that logit $\boldsymbol{h}$ is from the source domain. Similarly, the target encoder  can be trained in an adversarial setting with a minimax loss function $L_T^{\mathrm{MAX}} = \mathbb{E}_{(\boldsymbol{h}_t) \sim \mathbb{H}_T}\log(1-p_{\mathrm{domain}}(y_{b}=1|\boldsymbol{h}_t)$
or an  inverted label loss function $L_T^\mathrm{INV}=   - \mathbb{E}_{(\boldsymbol{h}_t) \sim \mathbb{H}_T}\log(p_{\mathrm{domain}}(y_{b}=1|\boldsymbol{h}_t))$.

 The simplest extension of ADDA to a multi-class variant that leverages on source task knowledge  would be add another $K$-dimensional head $H_\mathrm{task}$ to the discriminator.   This additional head performs task classification, and  trains the discriminator to classify the source examples only based on their task labels $y_s \in \{1,\dots ,K\}$.  The setup is illustrated in Fig. \ref{fig:D_config}(b) and is analogous to the  DANN \cite{ganin2016domain} except we have separate domain encoders and we replace the gradient reversal layer with a discriminator and adversarial training.  Additionally, the base configuration  is also representative of a discriminative variant to the AC-GAN \cite{odena2017conditional}, which adds a second classification head in the discriminator for stabilizing GAN training. However, we note that unlike AC-GAN, we assume no target labels in the unsupervised adaptation setting, so we can only train the task classification head on source examples.  In order to simplify the expressions, we can write the discriminator loss function for two heads in terms of posteriors as:

\begin{equation}
\begin{aligned}
L_{D',H_{\{\mathrm{domain},\mathrm{task}\}}}^\mathrm{MULTI} = & - \mathbb{E}_{(\boldsymbol{h}_s, y_s) \sim \mathbb{H}_S}\log(p_{\mathrm{task}}(y_s|\boldsymbol{h}_s)) \\ & + L^\mathrm{ADDA}_{D',H_\mathrm{domain}}
  \label{eq_two_heads}
\end{aligned}
\end{equation}

\noindent where $p_{\mathrm{task}}(y_s=k|\boldsymbol{h}_s)$ is the posterior probability output by $H_\mathrm{task}(D'(\boldsymbol{h}_s))_{k}$  that source logit $\boldsymbol{h}_s$ is from class with label  $y_s=k$. The first term represents the cross entropy loss  with source task labels  and the remaining terms equate to $L_{D',H_\mathrm{domain}}^{\mathrm{ADDA}}$.  As we only train the domain head adversarially, the adversarial loss function for training the target encoder is simply $L_T^\mathrm{INV}$ or $L_T^\mathrm{MAX}$.\footnote{Another option would be to  train  $H_\mathrm{task}$ adversarially in addition to $H_\mathrm{domain}$ by alternately minimizing and maximizing a distribution metric between\ $H_\mathrm{task}(D'(\boldsymbol{h}_s))$ and $H_\mathrm{task}(D'(\boldsymbol{h}_t))$.  However, as there is no `target domain' class, this intuitively means that the only way for the discriminator to maximize the metric would be to introduce intra-class confusion within the target domain - thus leading to instability during training.}   

We  can rewrite (\ref{eq_two_heads}) as:

\begin{equation}
\begin{aligned}
&L_{D',H_{\{\mathrm{domain},\mathrm{task}\}}}^\mathrm{MULTI} = \\&  - \mathbb{E}_{(\boldsymbol{h}_s, y_s) \sim \mathbb{H}_S}\log(p_{\mathrm{task}}(y_s|\boldsymbol{h}_s).p_{\mathrm{domain}}(y_{b}=1|\boldsymbol{h}_s)) \\ &   - \mathbb{E}_{(\boldsymbol{h}_t) \sim \mathbb{H}_T}\log(1-p_{\mathrm{domain}}(y_{b}=1|\boldsymbol{h}_t))
\label{eq_two_heads_2}
\end{aligned}
\end{equation}

%

 As is evident from the first term in (\ref{eq_two_heads_2}), with two  heads we are effectively optimizing the likelihood of  the joint posterior distribution over the task classes and source domain, but treating source domain classification and task classification as independent events.     Notably, as we only have access to labels in the source domain, the task classifier is only trained on source domain examples. As such, we can improve generalization by removing the independence assumption and model with a single multi-task classification head $H_\mathrm{joint}$:

\begin{equation}
\begin{aligned}
& L^\mathrm{JOINT}_{D',H_\mathrm{joint}}= \\ & - \mathbb{E}_{(\boldsymbol{h}_s, y_s) \sim \mathbb{H}_S}\log(p_{\mathrm{joint}}(y_s,y_{b}=1|\boldsymbol{h}_s)) \\ &   - \mathbb{E}_{(\boldsymbol{h}_t) \sim \mathbb{H}_T}\log(1-p_{\mathrm{joint}}(y_{b}=1|\boldsymbol{h}_t)) =  \\ &    - \mathbb{E}_{(\boldsymbol{h}_s, y_s) \sim \mathbb{H}_S}\log(p_{\mathrm{joint}}(y_s|\boldsymbol{h}_s, y_b=1). p_{\mathrm{joint}}( y_b=1|\boldsymbol{h}_s)) \\&  - \mathbb{E}_{(\boldsymbol{h}_t) \sim \mathbb{H}_T}\log(1-p_{\mathrm{joint}}(y_{b}=1|\boldsymbol{h}_t))
\label{eq_single_head}
\end{aligned}
\end{equation}


We illustrate this joint formulation in Fig. \ref{fig:D_config}(c). Essentially, by directly  optimizing the  joint posterior distribution $p_{\mathrm{joint}}(y_s,y_b=1|\boldsymbol{h}_s)    $, we can now also implicitly model  a conditional dependency $p_{\mathrm{joint}}(y_s|\boldsymbol{h}_s,y_b=1)    $ for task classification  given the source domain. Furthermore, if we marginalize over the task labels $y_s$, we end up with the standard ADDA loss formulation  $L_{D',H_\mathrm{domain}}^{\mathrm{ADDA}}$, of  (\ref{eq_ADDA}). 

 As in our proposal, we can write (\ref{eq_single_head}) in terms of a single $K+1$ classification head with $K+1$ labels $y \in \{1,\dots,K+1\},$ where the first $K$ classes model the joint distribution over task classes and source domain and the $K+1$-th class models the distribution over the target domain:  

\begin{equation}
\begin{aligned}
L^\mathrm{JOINT}_{D',H_{\mathrm{joint}}}= & - \mathbb{E}_{(\boldsymbol{h}_s, y) \sim \mathbb{H}_S}\log(p_{\mathrm{joint}}(y ,y<K+1|\boldsymbol{h}_s)) \\ &   - \mathbb{E}_{(\boldsymbol{h}_t) \sim \mathbb{H}_T}\log(p_{\mathrm{joint}}(y=K+1| \boldsymbol{h}_t))
\\ =&  -\mathbb{E}_{(\boldsymbol{h}_s, y_s) \sim \mathbb{H}_S}{\log(p_{\mathrm{joint}}(y_s|\boldsymbol{h}_s))} \\ &   - \mathbb{E}_{(\boldsymbol{h}_t) \sim \mathbb{H}_T}\log(p_{\mathrm{joint}}(y=K+1| \boldsymbol{h}_t))
\end{aligned}
\label{eq_single_head_2}
\end{equation}

 With the above notation,   $p_{\mathrm{joint}}(y=K+1| \boldsymbol{h})=(1-p_{\mathrm{joint}}(y_{b}=1|\boldsymbol{h}))  $ and $p_{\mathrm{joint}}(y ,y<K+1|\boldsymbol{h}_s)=p_{\mathrm{joint}}(y_s,y_b=1|\boldsymbol{h}_s))$. Finally, we can rewrite (\ref{eq_single_head_2}) in terms of the discriminator $D(\boldsymbol{h})_k = p_{\mathrm{joint}}(y=k| \boldsymbol{h})  $ and $D = H_\mathrm{joint}(D')$, which gives us the loss function for  a discriminative variant to semi-supervised GANs \cite{odena2016semi, salimans2016improved}:

\begin{equation}
\begin{aligned}
{L}^\mathrm{JOINT}_D  = & - \mathbb{E}_{(\boldsymbol{h}_s, y_s) \sim \mathbb{H}_S}\sum^{K}_{k=1} 1_{[k=y_s]}\log(D(\boldsymbol{h}_s)_k) \\ &  - \mathbb{E}_{(\boldsymbol{h}_t) \sim \mathbb{H}_T}\log(D(\boldsymbol{h}_t)_{K+1})
\end{aligned}
\label{DFM}
\end{equation}

We denote the first expectation term in (\ref{DFM}) as ${L}_{D,s}^\mathrm{JOINT}$ and the second expectation term as ${L}_{D,t}^\mathrm{JOINT}$. The loss function in (\ref{DFM}) corresponds with the discriminator  loss functions utilized for training semi-supervised GANs; the first term is the cross entropy term on the source examples with the $K$ task-specific labels and the second term is the cross entropy term on the target logits with the $K+1$-th `target' label. Our proposed discriminator loss function $L_D^\mathrm{REC}$  in (\ref{D1}) also follows the same format, except we substitute logits $\boldsymbol{h}$   for noisy logits $\boldsymbol{\tilde{h}} \sim N(\boldsymbol{\tilde{h}}|\boldsymbol{h})$ and substitute the indicator function $1_{[k=y_s]}$   with the source encoder posteriors $\boldsymbol{p}_s=C_s(\boldsymbol{h}_s)$, thus emulating a denoising autoencoder in the first term.

\subsection{Analysis of   Target Encoder Loss Functions, $L_T$  } \label{sec:semivsprop}

  We now perform an extended analysis of   target encoder loss functions given a single discriminator head, by considering discriminative variants to semi-supervised GANs and comparing with our proposed formulation, $L^\mathrm{MMD}_T(\mathbb{P}_S \rightarrow \mathbb{Q}_T)$.   Semi-supervised GANs are typically trained adversarially with either a minimax or feature matching objective function \cite{odena2016semi, salimans2016improved}. The discriminative variant of the minimax objective $L_T^\mathrm{MAX}$  for training the target encoder corresponds to   maximizing  (\ref{DFM}). For feature matching, the target encoder is trained to minimize a L2 distance-based loss  on the averaged intermediate source and target activations $f(\boldsymbol{h})$ of the  discriminator:              
\begin{equation}
{L}_T^\mathrm{FEAT}=\left\Vert \mathbb{E}_{(\boldsymbol{h}_s, y_s) \sim \mathbb{H}_S}({f}(\boldsymbol{h}_s))-\mathbb{E}_{(\boldsymbol{h}_t) \sim \mathbb{H}_T}({f}(\boldsymbol{h}_t)) \right\Vert^2_2
\label{TF}
\end{equation}

We note the equivalence between ${L}_T^\mathrm{FEAT}$ and an  MMD loss function between discriminator posteriors, $L_T^\mathrm{MMD}(\mathbb{Q}_S \rightarrow \mathbb{Q}_T)$. Following the notation, in Section \ref{sec:step_2}, for  infinite dimensional feature  map $\phi : \mathcal{X} \rightarrow\ \mathcal{H}$, where $\mathcal{H}$ represents reproducing kernel hilbert space (RKHS), this can be expressed as:   

\begin{equation}
L_{T}^\mathrm{MMD}(\mathbb{Q}_S \rightarrow \mathbb{Q}_T)= \left\Vert \mathbb{E}_{\boldsymbol{q}_s \sim \mathbb{Q}_S}\phi(\boldsymbol{q}_s) - \mathbb{E}_{\boldsymbol{q}_t \sim \mathbb{Q}_T}\phi(\boldsymbol{q}_t) \right\Vert^2_\mathcal{H}
\label{MMD_Q}
\end{equation}

The difference between ${L}_T^\mathrm{FEAT}$ and $L_{T}^\mathrm{MMD}(\mathbb{Q}_S \rightarrow \mathbb{Q}_T)$ is that  ${L}_T^\mathrm{FEAT}$ aligns intermediate features in finite dimensional Euclidean space, whereas $L_{T}^\mathrm{MMD}(\mathbb{Q}_S \rightarrow \mathbb{Q}_T)$ aligns the projected discriminator posteriors
in infinite dimensional RKHS. Notably,  MMD employed in (\ref{MMD_Q}) (and in our proposal)  can  be interpreted as matching all moments between the source and target posterior distributions, whereas conventional feature matching of ($\ref{TF}$)  is only empirically matching the first order moments (means) of the intermediate discriminator layer activations. 

In order to transition from (\ref{MMD_Q}) to our proposed target encoder loss function $\mathcal{L}^\mathrm{MMD}_T(\mathbb{P}_S \rightarrow \mathbb{Q}_T)$, we simply replace the distribution over source discriminator posteriors $\mathbb{Q}_S$ with the distribution over source encoder posteriors $\mathbb{P}_S$. The problem with $\mathcal{L}^\mathrm{MMD}_T(\mathbb{Q}_S \rightarrow \mathbb{Q}_T)  $ (and feature matching) is that it suffers from \textit{internal covariate shift}\footnote{Internal covariate shift is  the phenomenon wherein the distribution of inputs to a layer
in the network changes due to an update of parameters of the previous layers, and is typically synonymous with batch normalization \cite{ioffe2015batch}, which tries to minimize internal covariate shift by normalizing each layer to be zero mean and unit variance.} ; this is because it is effectively aligning the distribution over target discriminator posteriors $\mathbb{Q}_T$ \textit{to a changing source reference} $\mathbb{Q}_S$, as both $\mathbb{Q}_S$ and $\mathbb{Q}_T$ are parameterized by the discriminator (which is being trained). The constantly changing  $\mathbb{Q}_S$ adds noise to the target encoder alignment and destabilizes training. On the contrary, in our proposal, $\mathcal{L}^\mathrm{MMD}_T(\mathbb{P}_S \rightarrow \mathbb{Q}_T)$,  we align $\mathbb{Q}_T$ to the distribution over the source encoder posteriors $\mathbb{P}_S$, \textit{which is fixed during adversarial training} and only changes stochastically with mini-batch, as the source encoder has already been trained with source labels\footnote{It is worth noting that both our discriminator and target encoder loss functions, $L_D^\mathrm{REC}$  and $L_T^\mathrm{MMD}(\mathbb{P}_S \rightarrow \mathbb{Q}_T)$ respectively,  are centralized on the fixed distribution $\mathbb{P}_S$}. In addition, the source encoder posteriors $\boldsymbol{p}_s \sim \mathbb{P}_S$ are $K$-dimensional, which we extend with a zero to represent the $K+1$-th `target domain' class probability. This zero constraint on the `target domain' class acts as a stronger prior, in which  we enforce that the target domain examples are coming from the source. The combination of the zero constraint and fixed source reference provides improved stability during alignment of the target to source distribution; this is reflected later in  Table \ref{tab:combinations}, where accuracy with $\mathcal{L}^\mathrm{MMD}_T(\mathbb{P}_S \rightarrow \mathbb{Q}_T)$ is shown to be  substantially higher than $\mathcal{L}^\mathrm{MMD}_T(\mathbb{Q}_S \rightarrow \mathbb{Q}_T)$.


For the sake of completeness, we propose a final discriminative variant for the target encoder loss function, inspired by  unsupervised GAN training, where the generator is commonly trained with an inverted label objective $\mathrm{L}_T^\mathrm{INV}$ (i.e., inverting the generator label and training with cross entropy). As the inverted label objective is not viable for a multi-class discriminator output in our proposal, we instead propose  a pseudo-label objective for training the target encoder in the discriminative setting. This objective    draws parallels to unsupervised  domain adaptation  work that use pseudo-labels (typically in conjunction with  co-training). The pseudo-label is taken as the index of the maximum of the first  $K$ discriminator logits  $\boldsymbol{h}_d$. In other words, denoting \(\hat{y}_t =   \operatorname*{argmax}_{{j \in 1,\dots,K }} \boldsymbol{h}_d\), we train the target encoder by minimizing:

\begin{equation}
{L}^\mathrm{PSEUDO}_{T} = {- \mathbb{E}_{(\boldsymbol{h}_t) \sim \mathbb{P}_T}\sum^{K}_{k=1}1_{[k=\hat{y}_t]}\log(D(\boldsymbol{h}_t)_k)}
\label{target_loss_1}
\end{equation}

We note that, unlike our proposed target encoder loss function that is distribution-based, both  inverted label assignment and  our pseudo-label assignment are  instance-based. This potentially  means they are more prone to instability from noisy examples in the training batch. 

In order to motivate our proposed  adversarial loss functions compared to these discriminative variants of semi-supervised GANs, we perform an extensive ablation analysis in Section \ref{sec:ablation} on the SVHN $\rightarrow$ MNIST domain adaptation task.

%


%
%
%
%
%
%
%

\begin{figure*}
\centering

\subfloat[SVHN $\rightarrow$ MNIST (source only) ]{\includegraphics[scale=0.35]{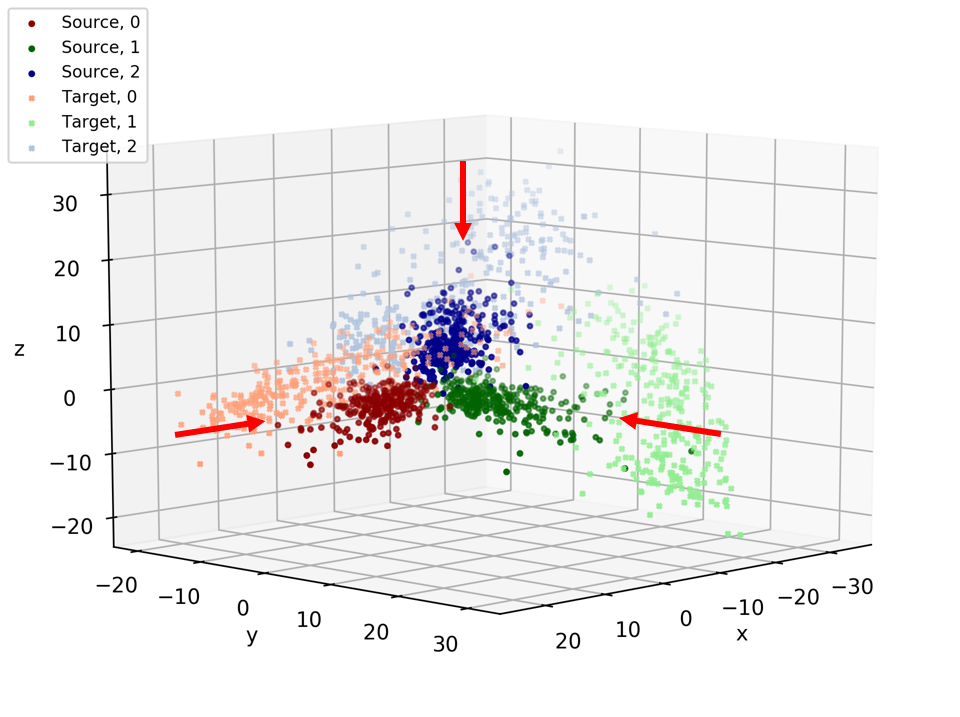}} \subfloat[SVHN $\rightarrow$ MNIST (contraction, no target reg.)  ]{\includegraphics[scale=0.35]{{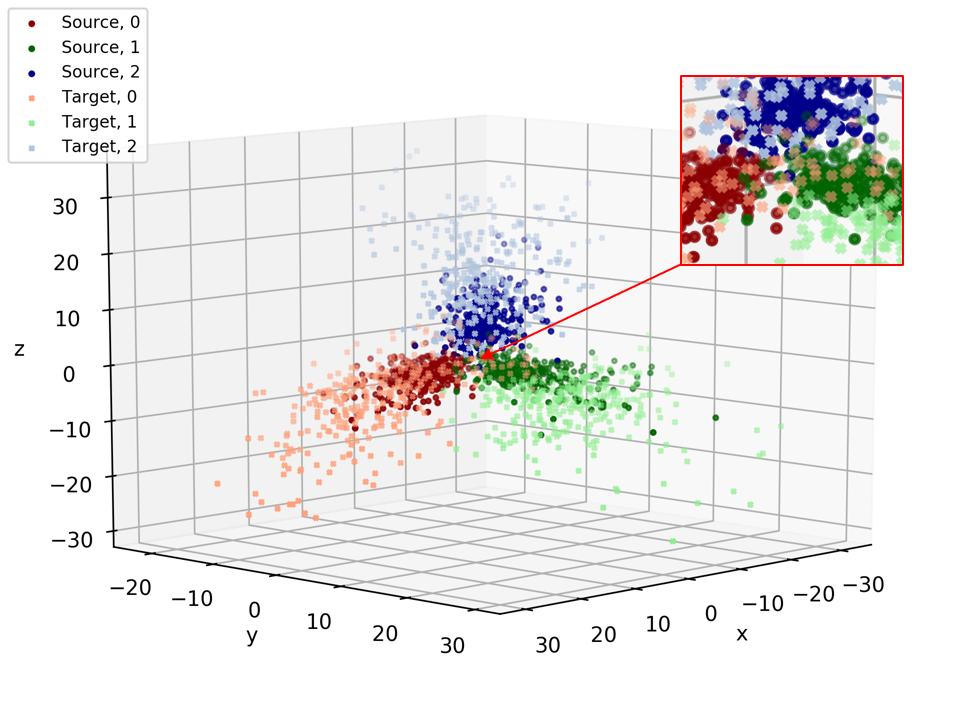}}} \subfloat[SVHN $\rightarrow$ MNIST (contraction, target reg.)  ]{\includegraphics[scale=0.35]{{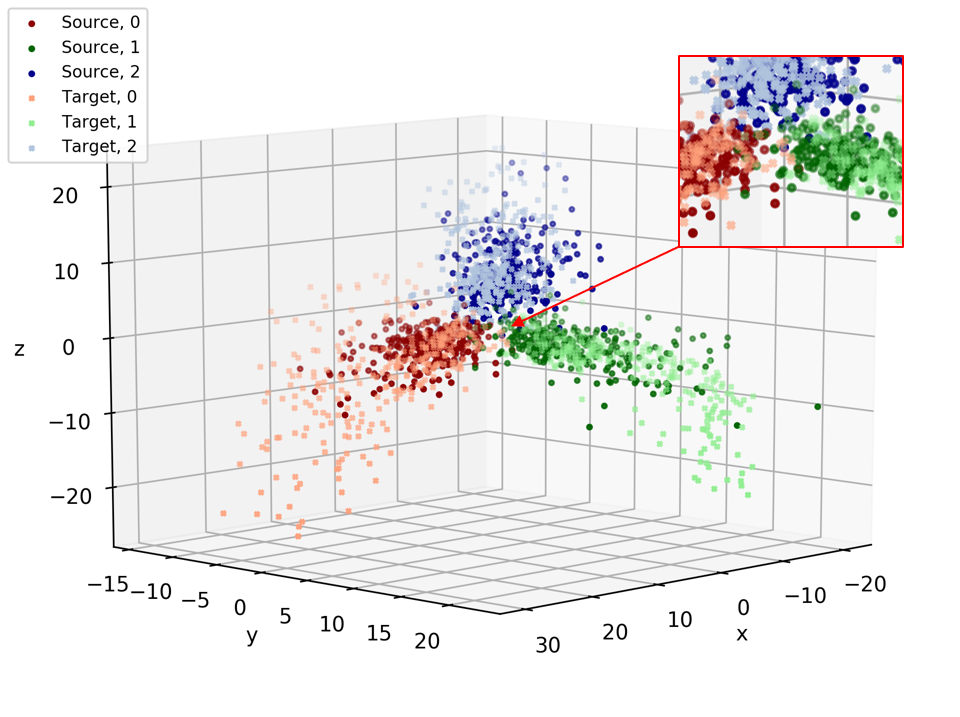}}}
\\
\subfloat[MNIST $\rightarrow$ MNISTM (source only)  ]{\includegraphics[scale=0.35]{{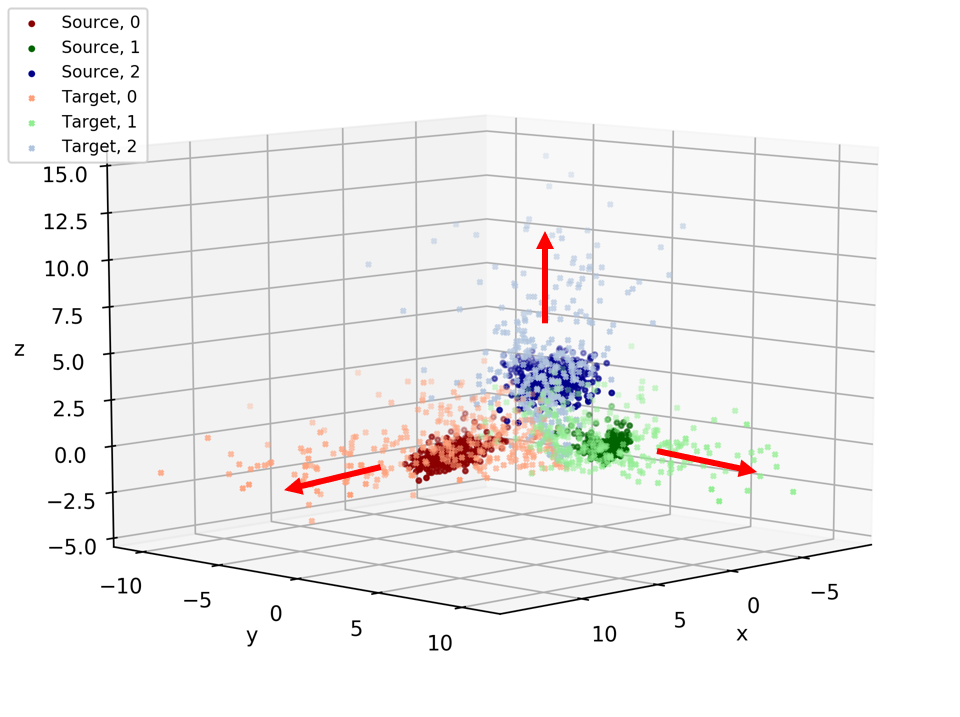}}} \subfloat[MNIST $\rightarrow$ MNISTM (expansion, no target reg.)  ]{\includegraphics[scale=0.35]{{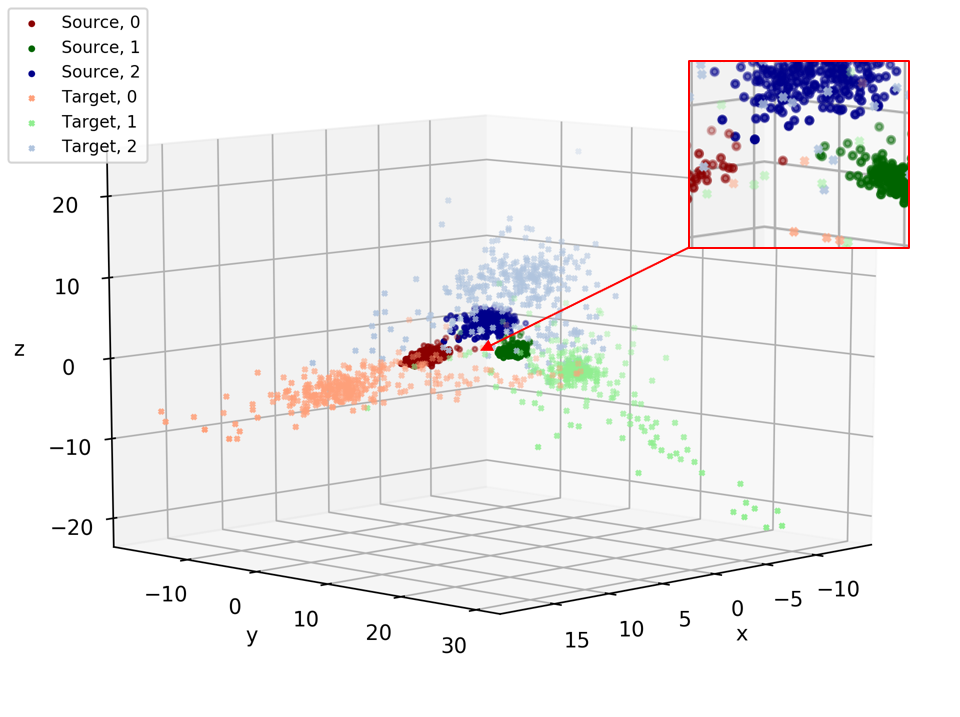}}}
\subfloat[MNIST $\rightarrow$ MNISTM (expansion, target reg.)  ]{\includegraphics[scale=0.35]{{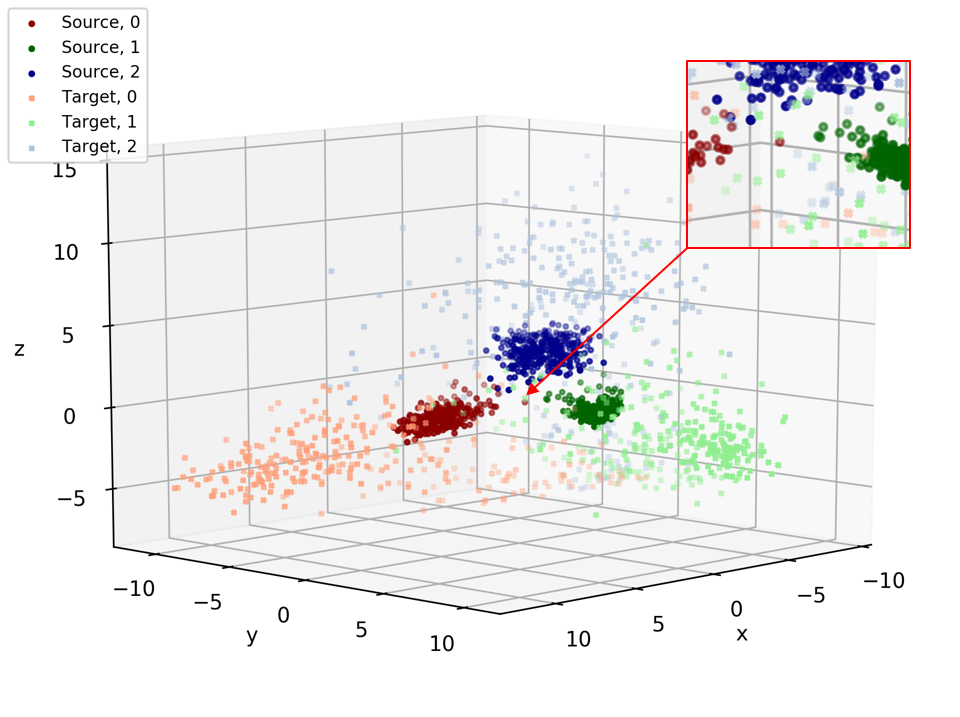}}}

\caption{(Best viewed in color) 3D scatter plot of source and target logits for  SVHN $\rightarrow$ MNIST and MNIST $\rightarrow$ MNISTM domain adaptation tasks on 3 classes only (0, 1 and 2). Pale points represent target logits and dark points represent source logits, color coded by class.   The plots represent the following: (a), (e): source pre-training only on target encoder; (b), (c): adversarial training with no target regularization; (c), (f): adversarial training with target regularization. The red arrows in (a) and (d) represent the direction of alignment  for    SVHN $\rightarrow$ MNIST    and MNIST $\rightarrow$ MNIST-M respectively during  adversarial training;  we refer to this  as a contraction mapping for SVHN $\rightarrow$ MNIST  (i.e., the target per-class distributions are being contracted in (b) and (c)) and an expansion mapping for MNIST $\rightarrow$ MNIST-M.  Source and target examples are randomly selected from the  test datasets respectively for visualization.}
\label{fig:disjoint}
\end{figure*} 

\section{Target Regularization with Source Examples} \label{sec:reg}

We can further improve on the convergence properties of adversarial training  by considering the various adaptation scenarios that arise     after initializing $E_t$ with $\theta_s$.  We illustrate respectively in the first and second rows of Fig. \ref{fig:disjoint},    contraction and expansion mappings on the 3 class SVHN $\rightarrow$ MNIST  and MNIST $\rightarrow$ MNISTM domain adaptation tasks, after initializing the target encoder with the source weights. Essentially, a contraction mapping is where the   target per-class distributions  must contract or converge towards the source origin during adversarial training, in order to align with the source per-class distributions.    Conversely, an expansion mapping is where the target per-class distributions must expand or diverge away from the source origin in order to align with the source per-class distributions. The origin represents a fixed point of uncertainty in classification (where the logits represent a uniform distribution).   

The target distribution undergoes a contraction mapping under adversarial training for  SVHN $\rightarrow$ MNIST, with the  contraction direction represented with red arrows in Fig. \ref{fig:disjoint}(a). For this contraction mapping, we note that during adversarial training, the target logits can overfit to the source and misclassify around the origin - this is evident in Fig. \ref{fig:disjoint}(b), where we show that target logits with label `0' (pale red points)  are being misclassified as `1'(pale green points) or `2' (pale blue points).  As adversarial alignment is only focused on aligning the entire distribution, it does not consider the class decision boundaries. This aspect is discussed in detail by Saito \textit{et al.} \cite{saito2018maximum}. However, if the source is well defined around its origin, we can regularize training of the target encoder with source examples, in order to minimize overfitting and negative transfer around the source origin.  In this way, we can enforce  better class separation around the origin.  Indeed, when comparing Fig. \ref{fig:disjoint}(b) (no target reg.) and (c) (with target reg.), we see that the decision boundaries between the 3 classes are more clearly defined with target regularization. 

The target distribution undergoes an expansion mapping under adversarial training for  MNIST $\rightarrow$ MNIST-M, with the expansion direction represented with red arrows in Fig. \ref{fig:disjoint}(d). For this expansion mapping,   regularizing with source examples instead has a  negative bias.  Essentially, whereas the target per-class distributions are diverging away from the source origin, the additional source examples bias the target distribution \textit{towards} the origin, thus reducing class separation.  While this bias may not be immediately clear when comparing  Figs. \ref{fig:disjoint}(e) (no target reg.) and (f) (with target reg.), it is apparent when comparing the axis scales that target regularization is constraining the support of the target distribution within the range [-10,10] for each axis.

We define the distribution $\mathbb{D}_{S \cup T}$ as the distribution of the union over the set of source and target examples $\mathbf{X}_{S \cup T} = \mathbf{X}_S \cup \mathbf{X}_T$. During adversarial training, each mini-batch to the target encoder is composed of 50\% source examples and 50\% target examples. We present results on the contraction  (SVHN $\rightarrow$ MNIST) and expansion (MNIST $\rightarrow$ MNISTM) setting in Table \ref{tab:target_reg} (as illustrated in Fig. \ref{fig:disjoint}).  As expected, while the regularization does improve results for SVHN $\rightarrow$ MNIST, there is a slight detrimental effect on MNIST $\rightarrow$ MNISTM as, in this case, adding source examples   has a negative bias as training progresses. 

A heuristic method for identifying whether the target distribution undergoes an expansion or contraction mapping during alignment would require computing the per-class statistics of the target distribution.   However, in the unsupervised domain adaptation setting, we assume no access to target labels and therefore can not compute the target per-class statistics. Therefore, in this paper, we instead propose to use a simple bagging process, by training  models with and without target regularization and performing a weighted averaging of logits during inference.
The weights are computed based on the L1 normalized maximum `confidence' in the predicted class (i.e., maximum value post-softmax). In this way, target regularization with source examples is used in this paper as the means to better generalize our proposal across multiple datasets. 
\begin{table} \centering \caption{Accuracy on SVHN $\rightarrow$ MNIST and MNIST $\rightarrow$ MNISTM task with 3 (0,1 and 2) classes for our proposed loss formulation with and without target regularization.}\begin{footnotesize} \resizebox{0.5\textwidth}{!} {\begin{tabular}{cccccccccccc} \toprule  & MNIST $\rightarrow$ MNISTM & SVHN $\rightarrow$ MNIST \\

  \midrule
Source only   & 0.798 & 0.795   \\

Target reg. & 0.866 & 0.980 \\

No target reg. & 0.905 & 0.948 \\

\bottomrule 
 \end{tabular}} \end{footnotesize}  \label{tab:target_reg} \end{table}

\section{Experimental Results and Analysis} \label{sec:exp_results}

We present experimental results   and analysis on the unsupervised domain adaptation task. In order to compare with ADDA and other recently proposed methods, we experiment on four digits datasets of varying sizes and difficulty: MNIST-M \cite{ganin2016domain}, MNIST \cite{lecun1998gradient},  USPS  and SVHN \cite{netzer2011reading}. We demonstrate substantial gain over ADDA and other recent methods, which is evident on the more difficult domain adaptation tasks such as SVHN $\rightarrow$ MNIST.   We additionally report accuracy on the  Office-31 dataset \cite{saenko2010adapting} compared to the current  state-of-the-art methods.
Finally, since neuromorphic vision
sensing presents a pertinent application for domain adaptation, we  introduce and validate
on a new NVS sign language dataset, demonstrating substantial gain in target accuracy compared
to training with the source domain only. For each domain adaptation task, we extract   5\%\ of each target adversarial  training split for validation, in order to tune the hyperparameters. To ensure consistency and demonstrate  lack of sensitivity to hyperparameters, we fix these globally over all tasks. Specifically, for the  MMD radial basis function (RBF)\ kernel combination $k$ (as described in Section \ref{sec:step_2}), we found that on average, the best performance to computational cost  for our framework is achieved with a summation over five kernels, with $\sigma_r = 10^{-r}, r \in \{0,\dots,4\}$.  Finally, as the discriminator is typically overcomplete (more nodes in the hidden layers than input classes), we add an L1 regularization term  to (\ref{D1}) on the discriminator weights $\boldsymbol{w}_d$ to improve the feature selection with regularization coefficient $\lambda=0.001$ for all cases.

 
\subsection{Digits datasets}

We consider four standard domain adaptation scenarios between dataset pairs drawn from MNIST-M \cite{ganin2016domain}, MNIST \cite{lecun1998gradient}, USPS and SVHN \cite{netzer2011reading} digits datasets, which are each comprised of $K=10$ digit classes (0-9).  Specifically, we evaluate on MNIST $\rightarrow$ USPS, USPS $\rightarrow$ MNIST, SVHN $\rightarrow$ MNIST and MNIST  $\rightarrow$ MNIST-M.  The difficulty in domain adaptation task increases as the variability between datasets increases.  We follow a similar training procedure to Tzeng \textit{et al.} \cite{tzeng2017adversarial}. For the MNIST $\rightarrow$ USPS and USPS $\rightarrow$ MNIST experiments, we sample 2000 images from MNIST and 1800 from USPS, otherwise we train and infer on the full datasets.  For MNIST $\rightarrow$ MNIST-M, we generate the unlabelled MNIST-M target dataset by following the process described by Ganin \textit{et al.} \cite{ganin2016domain}. We follow the architectures utilized by DANN \cite{ganin2016domain} and MCDDA \cite{saito2018maximum}.  Let us denote $\mathrm{Conv}(m,c)$ and $\mathrm{FC}(c)$ as convolutional  and fully connected layers respectively, with $m$ being the kernel size and  $c$  the  number of channels. Let us additionally denote pooling layers as $\mathrm{Pool}(w,s)$, where $w$ is the window size and $s$ is the stride.  Following this notation, our encoder architecture for larger datasets, (i.e., SVHN $\rightarrow$ MNIST, MNIST $\rightarrow$ MNIST-M) is $\mathrm{Conv}(5,64) \rightarrow \mathrm{Pool}(3,2) \rightarrow \mathrm{Conv}(5,64) \rightarrow \mathrm{Pool}(3,2) \rightarrow \mathrm{Conv}(5,128) \rightarrow \mathrm{FC}(3072) \rightarrow \mathrm{FC}(K)$ with discriminator $\mathrm{FC}(2048) \rightarrow \mathrm{FC}(K+1) $ (where $K$ is the number of task-specific classes).    We follow every convolutional layer with batch normalization and ReLU activation function, as per MCDDA. For smaller datasets (i.e., USPS $\rightarrow$ MNIST, MNIST $\rightarrow$ USPS), our encoder architecture is $\mathrm{Conv}(5,32) \rightarrow \mathrm{Pool}(2,2) \rightarrow \mathrm{Conv}(5,48)  \rightarrow \mathrm{FC}(100) \rightarrow \mathrm{FC}(K)$  with discriminator $\mathrm{FC}(500) \rightarrow \mathrm{FC}(500) \rightarrow \mathrm{FC}(K+1) $.    In step 1, the source encoder is trained with the Adam  optimizer \cite{kingma2014adam} for 10k iterations with a batch size of 128 and learning rate of 0.001.  In step 2, the target encoder is trained with a batch size of 128 per domain for 10k iterations,  with a lower learning rate of 0.0002, $\beta_1 = 0.5$ and $\beta_2 = 0.999$.  We resize all images to a fixed size of $28 \times 28$ prior to CNN processing. Additionally, we use data augmentation for  MNIST $\rightarrow$ MNIST-M   by randomly inverting the MNIST grayscale values and replicating the MNIST inputs channel-wise to match MNIST-M dimensions.  Our results when training on source only are provided in Table III. We also include results from several state-of-the-art methods as benchmarks, including   ADDA \cite{tzeng2017adversarial}, RAAN \cite{chen2018re}, DIFA \cite{volpi2018adversarial} and  MCDDA \cite{saito2018maximum},  which are recently proposed adversarial methods.

\ \subsubsection{Parametric exploration for discriminator loss function ${L}_D^\mathrm{REC}$}

\begin{figure}
\centering
\includegraphics[scale=0.12]{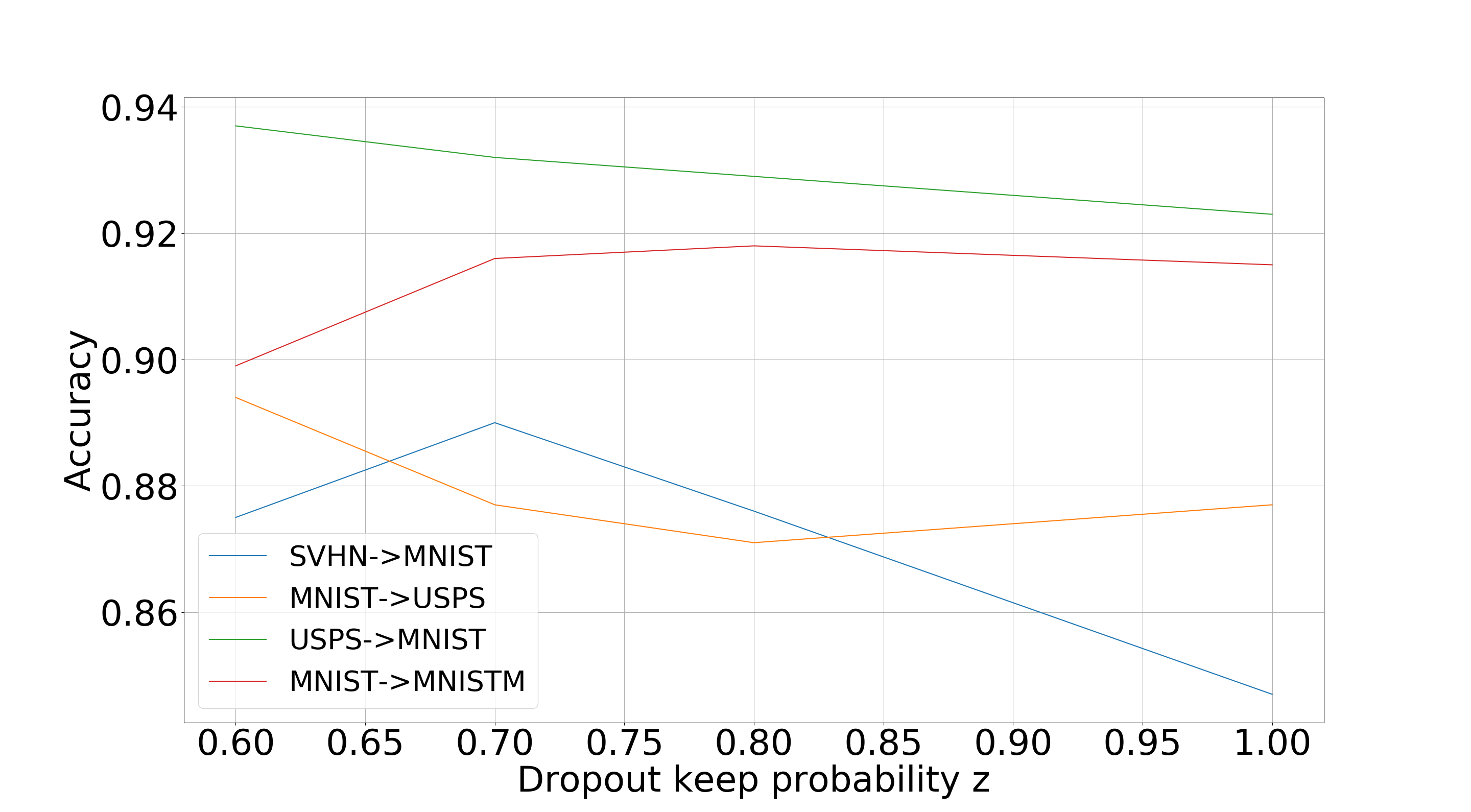}
\caption{Graph of dropout keep probability $z$ versus accuracy on subset of target dataset used for validation for various digits domain adaptation tasks (without target regularization).}
\label{fig:dropout_graph}
\end{figure}

 In Fig. \ref{fig:dropout_graph} we perform a parametric exploration over different values of dropout keep probability $z$ by computing the accuracy on the validation set for various digits domain adaptation tasks.     As discussed in Section \ref{sec:step_2}, adversarial training tends to destabilize when there are disjoint supports  between the target and source encoder distributions. With a dropout keep probability $z < 0.5$, over 50\% of classes will be randomly set to $0$ for the source and target  encoder logits,  $\boldsymbol{h}_s$ and $\boldsymbol{h}_t$ respectively.  It is now likely  that there is no  overlap between the remaining non-zero classes of the corrupted source and target logits.    In this case, the source and target encoder distributions (over logits) may not only be disjoint but lie in orthogonal spaces, which leads to a substantial drop in accuracy; for example, for SVHN $\rightarrow$ MNIST and $z=0.2$, the target accuracy attainable is only 51.3\%. Given this drop in accuracy, we only plot $z$ for $[0.6,1.0]$. No denoising corresponds to $z=1.0$. We note from the figure that including denoising either improves or maintains accuracy; in particular for the most difficult task SVHN$\rightarrow $MNIST, the accuracy improves by 5\% when decreasing $z$ from 1.0 to 0.7. As $z=0.7$ provides the most consistent gain for the four tasks, we fix $z$ to this value for the remainder of the paper.
\begin{figure*}
\centering
\subfloat[Proposed $\mathrm{MMD} (\mathbb{P}_s \rightarrow \mathbb{Q}_T) $ ]{\includegraphics[scale=0.35]{proposed_3_class_v2.png}} 
\subfloat[Minimax ($\mathrm{MAX}$)  ]{\includegraphics[scale=0.35]{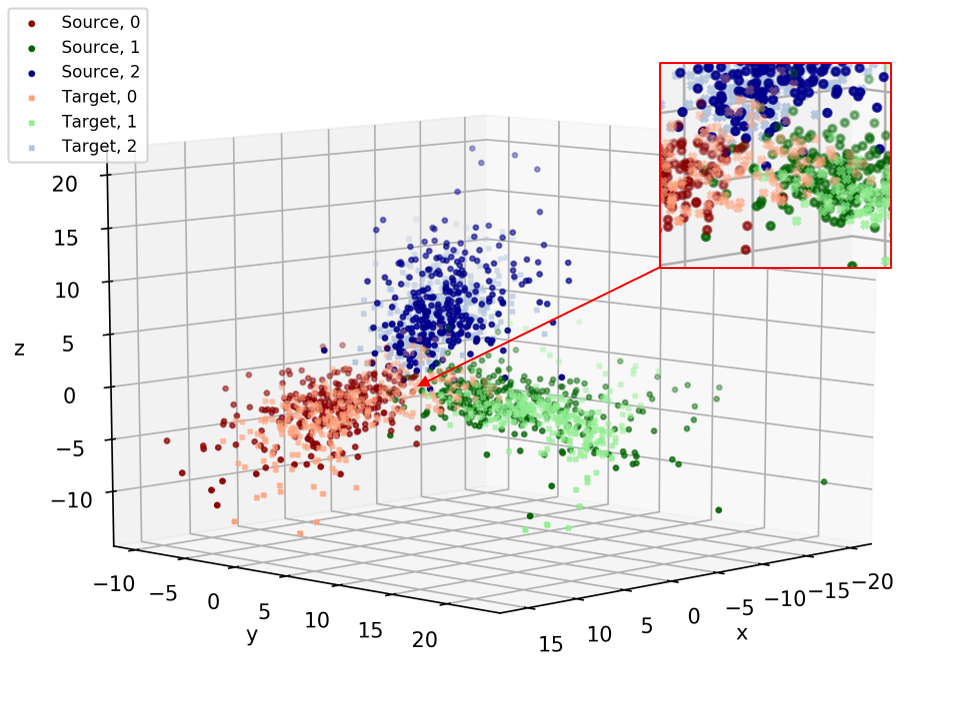}}  
\subfloat[Feature matching ($\mathrm{FEAT}$)]{\includegraphics[scale=0.35]{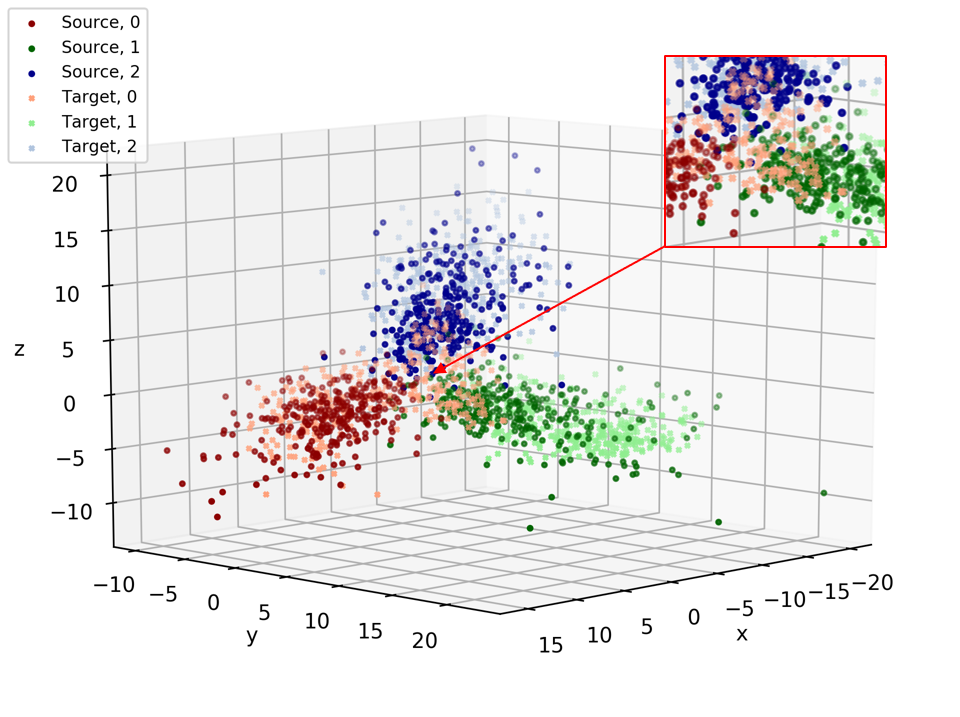}}
\caption{(Best viewed in color) 3D scatter plot of source and target logits for SVHN $\rightarrow$ MNIST domain adaptation task on 3 classes only (0, 1 and 2) for our proposed, minimax and feature matching loss formulations. Source and target examples are randomly selected from the SVHN and MNIST test datasets respectively for visualization.  \ The top-right of each scatter plot shows the zoomed section  around the source origin, where the distribution over classes is more uniform and  there is susceptibility to negative transfer. We see that decision boundaries at the origin are more clearly defined in (a) (our proposal) than in (b) or (c). }
\label{fig:loss_form}
\end{figure*}

\ \subsubsection{Ablation Study } \label{sec:ablation}

In order to illustrate the performance of our method and better understand where the performance gains are coming from, we perform an ablation analysis over various discriminator-encoder loss combinations  for the most difficult digit  domain adaptation task, SVHN$\rightarrow$MNIST. This includes the loss formulations introduced in Section \ref{sec:semi}.  Our results are presented in Table \ref{tab:combinations}, for all considered loss combinations, when training on 3 classes only (0,1 and 2) and all 10 classes. All hyperparameters and the training procedure are as described above for digits datasets; the only distinction is that for 3 classes we utilize our smaller dataset architecture (as utilized for USPS $\rightarrow$ MNIST) and only train on the source for 5000 iterations, in order to avoid overfitting.
 \\

\begin{table} \centering \caption{Accuracy on SVHN $\rightarrow$ MNIST task with 3 (0,1 and 2) and 10 classes for all considered discriminator (${L}_{D',H}$)-encoder (${L}_T$) loss combinations (as detailed in Section \ref{sec:semi}).  Each loss function is denoted by its corresponding column and superscript  (e.g., $\mathrm{ADDA} \rightarrow {L}_{D',H_\mathrm{domain}}^\mathrm{ADDA.}$).  $\mathbb{P}_S \rightarrow \mathbb{Q}_T$ refers to distribution alignment of  $\mathbb{Q}_T$ to $\mathbb{P}_S$.}  \resizebox{0.5\textwidth}{!} {\begin{tabular}{cccc} \hline ${L}_{D',H}$ & ${L}_T$  & 3 classes & 10 classes \\

\hline 
 \multirow{2}{*}{$\mathrm{ADDA}$}

& $\mathrm{INV}$ (Inverted label) &0.831& 0.787 \\
& $\mathrm{MAX}$ (Minimax) & 0.866 & 0.799 \\

\hline

 \multirow{2}{*}{$\mathrm{MULTI}$} 
 
 & $\mathrm{INV}$ (Inverted label) & 0.847& 0.783 \\
 & $\mathrm{MAX}$ (Minimax)  & 0.868& 0.796\\

 \hline

 \multirow{4}{*}{$\mathrm{JOINT}$}

&$ \mathrm{MAX}$  (Minimax) \cite{salimans2016improved, odena2016semi} & 0.856 & 0.753 \\

&$\mathrm{FEAT}$  (Feature matching)  \cite{salimans2016improved} & 0.805  & 0.772\\ 

&$\mathrm{PSEUDO}$  (Pseudo-label) & 0.863  & 0.800 \\

&$ \mathrm{MMD}$  $\mathbb{(P}_S \rightarrow \mathbb{Q}_T$)   & 0.895 & 0.856  \\

  \hline
  \multirow{2}{*}{$\mathrm{REC}$ (proposed)} 
&$\mathrm{MMD}$ ($\mathbb{Q}_S \rightarrow \mathbb{Q}_T$) & 0.878 & 0.804 \\  
  
& $\mathrm{MMD}$   ($\mathbb{P}_S \rightarrow \mathbb{Q}_T$) (proposed)   & \textbf{0.948} & \textbf{0.918}   \\

\hline 
 \end{tabular}}   \label{tab:combinations} \end{table}

\noindent\textbf{One head versus two heads: }  In the first two parts of Table \ref{tab:combinations}, we evaluate  performance of training   with ADDA  versus a multi-class variant of ADDA ($\mathrm{MULTI}$) with two classification heads (as introduced in Section \ref{sec:onevstwo_heads}). For these two benchmarks, we train the target encoder with the inverted label setting or minimax, i.e., $L^\mathrm{INV}_T$ and $L^\mathrm{MAX}_T$ respectively.  The results show  that adding the additional task classification head $H_\mathrm{task}$  in the $\mathrm{MULTI}$ configuration has little effect on accuracy compared to baseline ADDA.  We attribute this to the fact that the inputs $\boldsymbol{h}_s$  and outputs  of    $H_\mathrm{task}(D'(\boldsymbol{h}_s))$ are both learned with the  source labels, and the output  is not conditioned on the domain. The task classification head $H_\mathrm{task}$ simply  learns to invert the mapping learned by the preceding discriminator layers $D'$; $H_\mathrm{task} \sim {D'}^{-1}$.
This further  motivates our proposed learning with a single classification head that models the joint distribution between the task and source domain classification, as defined in (\ref{DFM}).   However, the third part of   Table \ref{tab:combinations} shows that there  appears to be a stronger discriminator bias and a slight detriment in accuracy for  $\mathrm{JOINT}-\mathrm{MAX}$ compared to the baseline $\mathrm{ADDA}-\mathrm{MAX}$  discriminator-encoder loss combinations. This motivates the need for our proposed encoder loss function $L_T^\mathrm{MMD}$.

\begin{table*} \centering \caption{Accuracy for  our proposed method  compared to the current state-of-the-art.       In order to isolate the performance gain from domain adaptation  for our proposals, we report in parentheses the percentage increase (relative) over the source-only accuracy,  as reported in the respective papers for DIFA \cite{volpi2018adversarial} and ADDA \cite{tzeng2017adversarial}.*UNIT \cite{liu2017unsupervised} and DTN \cite{taigman2016unsupervised}   use additional SVHN data (131 images and 531 images respectively).  **This is our implementation of ADDA \cite{tzeng2017adversarial} on MNIST $\rightarrow$ MNIST-M, as this task is not used in the original paper.  }\resizebox{0.85\textwidth}{!} {\begin{tabular}{cccccc} \toprule \multicolumn{2}{c}{Method}  & SVHN $\rightarrow$ MNIST & USPS $\rightarrow$ MNIST & MNIST $\rightarrow$ USPS  & MNIST $\rightarrow$ MNIST-M \\

  \midrule
\multicolumn{2}{c}{Source only} & 0.644 & 0.597   & 0.754  & 0.705  \\
 \midrule
 
\multicolumn{2}{c}{LSC \cite{hou2016unsupervised}} & - & 0.655 & 0.723 & -  \\ 

\multicolumn{2}{c}{NLSDT \cite{zhang2016lsdt}} & - & 0.791 & 0.874 & -  \\

\multicolumn{2}{c}{DANN \cite{ganin2016domain}} & 0.739 & 0.730& 0.771& 0.529\\

\multicolumn{2}{c}{DICD \cite{li2018domain}} & - & 0.652& 0.788 & - \\
 
\multicolumn{2}{c}{DDC \cite{tzeng2014deep}} & 0.681 & 0.665 & 0.791 & - \\

\multicolumn{2}{c}{DSN \cite{bousmalis2016domain}} & 0.827 & - & - & 0.832 \\

\multicolumn{2}{c}{DTN \cite{taigman2016unsupervised}} & 0.844* & -  & - & -  \\
\multicolumn{2}{c}{UNIT \cite{liu2017unsupervised}} & 0.905* &- & - & -  \\  
  
\multicolumn{2}{c}{CoGAN \cite{liu2016coupled}} &no convergence & 0.891 & 0.912 & - \\ 

\multicolumn{2}{c}{RAAN \cite{chen2018re}} & 0.892 & 0.921 & 0.890 & \textbf{0.985} \\

\multicolumn{2}{c}{ADDA \cite{tzeng2017adversarial}} & 0.760 (26\%) & 0.901 (58\%)& 0.894 (19\%) & 0.800 (14\%)**  \\
 
\multicolumn{2}{c}{DIFA \cite{volpi2018adversarial}} &  0.897 (32\%)  & 0.897 (43\%)& 0.923 (28\%) & -\\

\multicolumn{2}{c}{MCDDA \cite{saito2018maximum}} &  0.962 (43\%)  & 0.941 (48\%)& {0.942 (23\%)} & - \\ 
 
\multicolumn{2}{c}{PFAN \cite{chen2019progressive}} & 0.939 (56\%) & -& \textbf{0.950} \textbf{(26\%)} & - \\

\multicolumn{2}{c}{TPN \cite{pan2019transferrable}} & 0.930 (55\%) & 0.941 (60\%) & 0.921 (22\%)& -  \\ 
 
 \midrule 

 \multirow{3}{*}{Proposed}
& no target reg. & 0.918 (43\%) & 0.941 (58\%) & 0.895 (19\%) & 0.962 (36\%) \\

& target reg. &  \textbf{0.972 (51\%)} & \textbf{0.967 (62\%)} & 0.928 (23\%)& 0.955 (35\%) \\

& averaged & {0.964 (50\%)} & 0.966 (62\%) & 0.925 (23\%)& 0.960 (36\%) \\
\bottomrule

 \end{tabular}}  \label{tab:digits_stat} \end{table*}

\noindent\textbf{Proposed $L_{D}^\mathrm{REC},L_T^\mathrm{MMD}$  versus discriminative variants of semi-supervised GANs:} 
 We next consider how our proposed MMD based target encoder loss function ${L}_T^\mathrm{MMD}(\mathbb{P}_S \rightarrow \mathbb{Q}_T)$, improves over  conventional minimax  ${L}_{T}^\mathrm{MAX}$, feature matching  ${L}_{T}^\mathrm{FEAT}$ and pseudo-label  ${L}_{T}^\mathrm{PSEUDO}$ encoder loss formulations, as defined in Section \ref{sec:semi}. In order to isolate the performance of our proposed MMD\ loss function and perform a fair comparison, we fix the discriminator loss function to ${L}_D^\mathrm{JOINT}$. The third part of Table \ref{tab:combinations} shows that, on both 3 and 10 classes our proposed MMD\ loss formulation outperforms all other target encoder loss variants. In particular, when compared to feature matching, our proposal provides accuracy gains of over 11\%  on both 3 and 10 classes. In order to establish the source of this gain, we present 3D scatter plots in Fig. \ref{fig:loss_form} of the source and target logits when trained on 3 classes only from the SVHN $\rightarrow$ MNIST domain adaptation task.\footnote{We opted for this approach instead of using a reduction method such as t-SNE \cite{maaten2008visualizing} that introduces additional hyperparameters such as perplexity to visualize the domain shift.} As shown in the figure, both minimax and feature matching are prone to overfitting on the source dataset. While both formulations result in a tight bound on the source distribution, they forego a good class separation close to the source origin (as represented in top-right corner of each plot),   where the distribution over classes is more uniform and the target encoder loss would be smaller in magnitude, potentially unfavourably biasing towards the discriminator. There is negative transfer around the origin, which is most noticeable for the `0' digit class (pale red points), which is misclassified as `1' (green) or `2' (blue), and is worst for feature matching. This corresponds to the  accuracies reported in Table \ref{tab:combinations}, where feature matching is shown to perform the worst on 3 classes.    

 Having validated the performance gain from our proposed target encoder loss function, we now switch the discriminator loss function from ${L}_D^\mathrm{JOINT}$ to our proposed  reconstruction loss ${L}_D^\mathrm{REC}$ , with $z=0.7$. Combining our two proposed loss formulations for adversarial training,  ${L}_D^\mathrm{REC}$ and ${L}_T^\mathrm{MMD}(\mathbb{P}_S \rightarrow \mathbb{Q}_T)$, we are able to achieve the best performance on 3 and 10 classes. This is due to a combination of increased separability between the source and target domains  in RKHS, and the fixed source distribution and hard zero constraint on the `target' class that minimizes the internal covariate shift when training to align the distribution over target discriminator posteriors $\mathbb{Q}_T$. In order to isolate the detriment from internal covariate shift, we also consider aligning the distributions over source and target discriminator posteriors $(\mathbb{Q}_S \rightarrow \mathbb{Q}_T)$ in ${L}_T^\mathrm{MMD}$. The penultimate row of the table shows the performance if we align $\mathbb{Q}_T$ to $\mathbb{Q}_S$ in our encoder loss function; our accuracy drops substantially to 87.8\%\ and 80.4\%\ for 3 and 10 classes respectively, which illustrates the effect of the internal covariate shift, as in feature matching.\footnote{Whilst feature matching also suffers from internal covariate shift, it is also only aligning the empirical means between distributions, as discussed in Section \ref{sec:semi}.}

\subsubsection{Adaptation on digits datasets}

Finally, we report accuracy  for our proposed models with and without target regularization with source examples  (as discussed in Section \ref{sec:reg}) and after averaging the model logits. In order to isolate the performance gain from domain adaptation for the most competitive methods, we compute the percentage increase (relative) over the source only accuracy reported in the paper (shown in parentheses in Table \ref{tab:digits_stat}). On average, our proposal provides the best balance in performance over all datasets. The combination of target regularization and our proposed loss formulations lead to an accuracy of 97.2\% and 96.7\% on SVHN $\rightarrow$ MNIST  and USPS $\rightarrow$ MNIST respectively, surpassing all other methods and surpassing the recently proposed MCDDA by up to 2.6\%. (with a 51\% percentage increase over source training only on SVHN $\rightarrow$ MNIST).  Our method also outperforms recent prototypical network-based alignment methods, TPN \cite{pan2019transferrable} and PFAN \cite{chen2019progressive}, on the same datasets.\ However, we do note the one case where adding target regularization has a negative effect is  MNIST $\rightarrow$ MNISTM; it is most likely that this is an instance of expansion mapping, as discussed in Section \ref{sec:reg}, but on 10 classes.
Nonetheless, by performing a weighted averaging between  the logits of models with and without target regularization (bottom row of Table \ref{tab:digits_stat}), with the weights defined by the maximum `confidence' in the predicted class, we are able to provide good generalization  and improvement over the default setting of no target regularization for all datasets. While RAAN does perform 2.5\% better on MNIST $\rightarrow$ MNISTM, we note that the training process is  more complex, with 10 times the number of training iterations as in our implementation, and their final model is averaged over many different parameter selections and runs.

%

\subsection{Office-31 dataset}

\begin{table} \centering \caption{Accuracy for proposed configurations, when averaging over models with and without target regularization, compared to state-of-the-art on the Office-31 dataset.            } \resizebox{0.45\textwidth}{!} {\begin{tabular}{cccc} \toprule Method  & A $\rightarrow$ W & A $\rightarrow$ D & D $\rightarrow$ A   \\

  \midrule
   Source only & 0.707 & 0.720   & 0.581  \\
 \midrule

DASH-N \cite{nguyen2015dash} & 0.606 & - & -  \\

DANN \cite{ganin2016domain} & 0.730 & 0.723& 0.534\\
 
DDC \cite{tzeng2014deep} & 0.618 & 0.644 & 0.521 \\

DRCN \cite{ghifary2016deep} & 0.687  & 0.668 & 0.560  \\

  JAN \cite{long2016deep}  & 0.752 & 0.728  & 0.575 \\

ADDA \cite{tzeng2017adversarial} & 0.751  & - & -   \\

LDADA \cite{lu2018embarrassingly} & 0.781 & 0.767 & \textbf{0.683} \\

\midrule

Proposed, averaged & \textbf{0.836} &\textbf{ 0.809} & 0.622 \\

%
\bottomrule

 \end{tabular}}  \label{tab:office_stat} \end{table}     

 We report results on the standard Office-31 \cite{saenko2010adapting} dataset  in Table \ref{tab:office_stat}.   The Office-31 dataset consists
of 4,110 images spread across 31 classes in 3 domains:
Amazon,
Webcam, and
DSLR.  Our results focus on the three of the more difficult  domain adaptation tasks; Amazon $\rightarrow$ Webcam (A $\rightarrow$ W),  Amazon $\rightarrow$ DSLR (A $\rightarrow$ D) and DSLR $\rightarrow$ Amazon (D $\rightarrow$ A).  In order to demonstrate the strength of our proposal, we use VGG-16 pre-trained on ImageNet and fine-tune only the final fully-connected layer.
We train with stochastic gradient descent and a learning rate of 0.001. We set the dropout keep probability $z=0.7$ as in the digits task. Our discriminator is restricted to only 500 hidden units per layer and we only train adversarially for 2k iterations.  We note that the number of training parameters is  377 thousand in total, compared to over 6 million utilized for ADDA \cite{tzeng2017adversarial}. Despite only training on a small subset of total parameters, our proposal remains competitive or surpasses the performance of other recent methods.  We additionally note that under our training setup, ADDA consistently obtains a degenerate solution due to instability during training. 

We note that LDADA \cite{lu2018embarrassingly} outperforms our method on D $\rightarrow$ A. We attribute this to the insufficient amount of source training data available for this adaptation task.  The source dataset, DSLR, is only comprised of 498 images, compared to the target dataset, Amazon, which is comprised of 2817 images. Our framework tends to perform better when the source examples form higher density clusters, which the target examples can be aligned towards in RKHS - with only 498 images (on average 16 examples per class) in DSLR, we are unable to learn a discriminative distribution over source encoder posteriors  and therefore our model is outperformed for this particular task by LDADA.  However, while LDADA does perform well with limited source training data, it does also suffer from negative transfer, as mentioned in the paper - that is; when the domain discrepancy is small, LDADA degrades the accuracy during iterative validation, such that target examples are unavoidably misclassified.   This is likely why LDADA performs worse than our method on other domain adaptation tasks such as  A $\rightarrow$ W, where the domain discrepancy after source pre-training is noticeably smaller.


\subsection{NVS ASL dataset}

\begin{figure*}
\centering
{\includegraphics[scale=0.6]{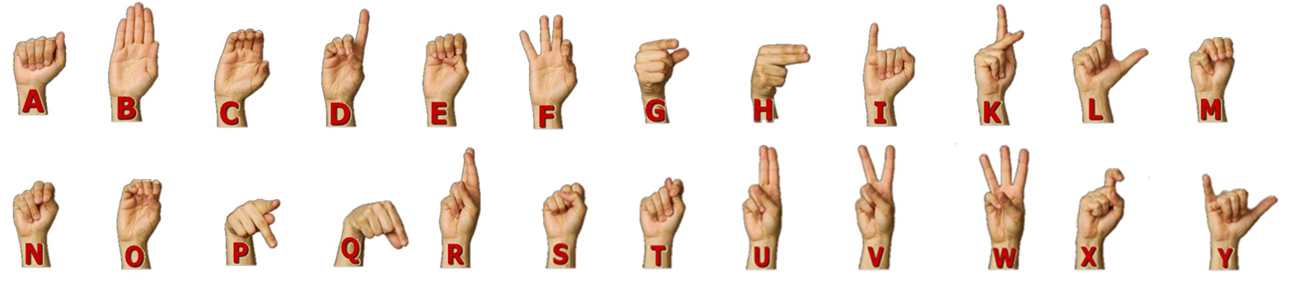}}
\caption{Signs for letters A-Z from the American Sign Language dataset.  Note that some letters such as M and N only have subtle differences. Letters J and Z are excluded given that they are not static signs and require a particular gesture.  }
\label{fig:asl}
\end{figure*}

\begin{figure*}
\centering
\subfloat[Emulated (source domain)]{\includegraphics[scale=0.4]{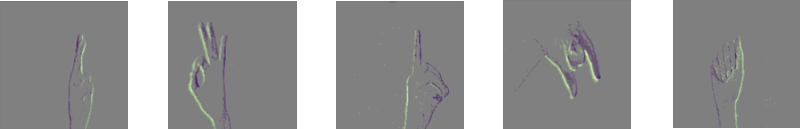}} \hspace{15pt}
\subfloat[Real (target domain)]{\includegraphics[scale=0.4]{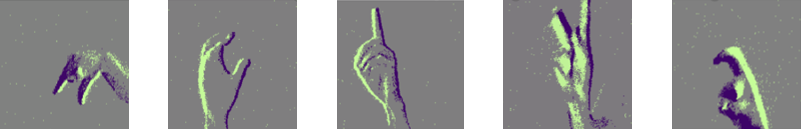}}
\caption{(Best viewed in color) Select frames from the NVS American Sign Language recognition dataset for a) emulated NVS frames (source domain) and b)  real NVS camera frames (target domain). The green/purple points correspond to the +1/-1 (or ON/OFF) spike polarity.  }
\label{fig:nvs_domain}
\end{figure*}

\begin{table*}[t] \centering \caption{Overall and per-letter recognition accuracy for select letters  of the NVS ASL dataset.  We evaluate on the source-only and our proposed method, when averaging over models with and without target regularization.     } \begin{footnotesize}\resizebox{0.9\textwidth}{!} {\begin{tabular}{cccccccccccc} \toprule Letter & A & B & E & F & M & N & S & T & U & V & overall \\

  \midrule
   Source only & 0.925 & 0.745   & 0.809 & 0.163  & 0.164 & 0.376 & 0.837 & 0.162 & 0.277 & 0.992 & 0.638 \\
\midrule
 ADDA \cite{tzeng2017adversarial} &  {0.988}&  {0.962}&  {0.261}&  {0.865}&  {0.152}&   {\textbf{0.808}}&   {{0.810}}&   {0.562}&   {0.954}&   \textbf{1.000}&   {0.837}\\
   MCDDA \cite{saito2018maximum} &  {0.988}&  {0.954}&  {0.010}&  {0.923}&  {\textbf{0.693}}&   {0.642}&   \textbf{0.893}&   {0.107}&   {\textbf{0.961}}&   \textbf{1.000}&   {0.793}\\
   DIFA \cite{volpi2018adversarial} &  {0.974}&  {0.958}&  \textbf{0.912}&  {0.884}&  {0.373}&   {0.571}&   {0.726}&   {0.631}&   {0.937}&   \textbf{1.000}&   {0.854}\\
\midrule
   $\mathrm{MULTI}-\mathrm{MAX}$ &  {0.988}&  {0.962}&  {0.868}&  {0.869}&  {0.361}&   {0.646}&   {0.770}&   {0.643}&   {0.942}&   \textbf{1.000}&   {0.850}\\
   $\mathrm{JOINT}-\mathrm{PSEUDO}$ &  {0.964}&  {0.954}&  {0.549}&  {0.919}&  {0.066}&   {0.469}&   {0.740}&   {0.445}&   {0.959}&   {0.996}&   {0.759}\\
   $\mathrm{REC}-\mathrm{MMD}(\mathbb{Q}_S \rightarrow \mathbb{Q}_T)$ &  {0.981}&  \textbf{0.970}&  {0.814}&  {0.880}&  {0.212}&   {0.571}&   {0.189}&   {0.593}&   {0.949}&   {0.996}&   {0.823}\\

\midrule
%
%

Proposed, averaged & \textbf{1.000}  & 0.966 & {0.895} & \textbf{0.938} & 0.361 & {0.699} & 0.818 & \textbf{0.650} & {0.953} & \textbf{1.000} & \textbf{0.873}\\

\bottomrule

 \end{tabular}} \end{footnotesize}  \label{tab:accuracy_stat} \end{table*}

We introduce a new sign language recognition dataset for NVS-based unsupervised domain adaptation.   The primary motivation behind creating the dataset and validating our framework with it is that progress in neuromorphic spike-based event or action recognition is severely hampered from the lack of NVS training data with reliable annotations \cite{tan2015benchmarking}. This is partially addressed via  emulators, which convert annotated APS\ video datasets into emulated NVS data in order to train advanced discriminative models in a supervised manner. However, beyond the unavoidable  gap between the experimental and the emulated NVS\ data distributions, the NVS\ camera technology is in constant evolution and new versions of hardware devices like DAVIS\ and ATIS\ \cite{delbruck2016neuromorphic}  and their multiple settings cause further domain shift against their previous versions and previously-released software emulation frameworks.

Our  experimental dataset is comprised of 1200 unlabelled real recordings and 1200 labelled emulated recordings, each representing a different static sign of 24 letters (A-Y, excluding J) from the American Sign Language (ASL).  We note that similar to other APS-based sign language recognition tasks \cite{pugeault2011spelling}, letters J and Z are excluded as their ASL designation requires motion. Fig. \ref{fig:asl} shows the required hand pose for each letter of the dataset.  As is evident from the figure, sign language recognition  presents a substantially more difficult task than digit recognition, considering that for  some letters (e.g., M and N) there is very little variation in fingers' positioning.


In order to generate the emulated spike events we first record  APS\ video of someone performing the sign for each letter  with  translational and rotational motion over the video duration, thus increase the difficulty of the recognition task.  Next, the APS\ video is recorded with a standard laptop camera, and consecutive APS frames are passed into the PIX2NVS\ emulator. PIX2NVS converts the APS frames to the  corresponding  emulated NVS frames, and this constitutes our source domain. The real NVS recordings are recorded directly with an iniLabs DAVIS240c NVS camera, again with rotational and translational motion over the video duration.

Fig. \ref{fig:nvs_domain} shows a selection of emulated  and real NVS frames  from the dataset.  There is a discernible  domain shift between the emulated and real spike events, with the real NVS events exhibiting a substantially higher spike density that increases the visibility of the signed letter, despite  also carrying  some background     `salt \& pepper' noise.  Nonetheless, we are able to demonstrate that our proposed method can reduce this domain shift.  As the recordings represent static signs we train on individual frames and remove a subset of frames from the start and end of the recording  where there may be no sign distinguishable. As such, we have $\sim 80,000$ emulated NVS frames for source training and $\sim 50,000$ real NVS frames.  We use $\sim 40000$ of the real NVS frames for domain adaptation and $\sim 10000$ for inference.\      The frame resolution in both domains is $240 \times 180$. Our source  encoder is VGG-16 \cite{simonyan2014very}, which we   train in step 1 on the emulated NVS frames  using stochastic gradient
descent with momentum set to 0.9. The learning rate is
set to 0.001, the batch size to 24 and we complete
training at 15k iterations. In terms of data augmentation, we first resize the input such that the smaller side is 256 and keep the aspect
ratio. We then use a multi-scale random cropping of the resized
RGB frame; the cropped volume is subsequently randomly
flipped, and normalized according to its mean.  In step 2, we initialize the target encoder from the source pre-trained weights and follow the same procedure with  data augmentation on both input domains, but   only train the target encoder fully connected layers adversarially for 10k iterations and fix all convolutional layers.   Contrary to the APS datasets, the discriminator is again restricted  to 500 hidden units per layer. We infer on the target dataset by  extracting a single center crop.

We present both the  overall and per-letter recognition accuracy in Table \ref{tab:accuracy_stat} when evaluating on the NVS ASL dataset. For clarity, we only include letters in the table with  subtle differences in sign configuration such as  M and N.  We include results on our proposed framework (averaged with and without target regularization), source training only, as well as ADDA and other recent competing methods. \ For recent work, we follow a similar  architecture/training procedure as in our proposal, to ensure  a fair test.  For the case of MCDDA, we follow the framework in \cite{saito2018maximum}.  In order to keep training/inference  complexity of the same order as our proposal and other methods, we constrain each of the two  classifiers to the final two fully connected layers of VGG-16 and the feature generator to all other VGG-16 layers. Nonetheless, we find that MCDDA\ is extremely sensitive to classifier/generator imbalance, which is presumably exacerbated by the sparsity of the NVS data  - overall it performs $4\%$  worse than ADDA. On the other hand, our proposal provides substantial increase in accuracy compared to training on the source only, and also outperforms ADDA on most letters and overall, by 3.6\%.  
By looking at the per-letter accuracies we can distinguish where ADDA substantially underperforms compared to our proposal; namely, on the set $\bm{C}$ = \{E,M,N,S,T\}.  If we cross-reference with Fig. \ref{fig:asl}, we note that the letters in this set are  not easily distinguishable from each other, which would be made worse when transforming to the NVS domain. Whereas ADDA generally performs poorly on all letters in the set, and effectively misclassifies the majority of instances of the letters $E$ and $M$ in order to align the target to the source  domain, our proposal is able to better transfer some of  the class separation learned from training on the source domain with labels and maintains consistently good accuracy across the set. 

We also evaluate on the NVS ASL dataset with the best performing variants (outside of our proposal) from Table \ref{tab:combinations}; namely, $\mathrm{MULTI}-\mathrm{MAX}$, $\mathrm{JOINT}-\mathrm{PSEUDO}$ and $\mathrm{REC}-\mathrm{MMD}(\mathbb{Q}_S \rightarrow \mathbb{Q}_T)$. As expected, the additional classification head and the minimax objective for $\mathrm{MULTI}-\mathrm{MAX}$ results in slightly improved performance over standard  ADDA trained with the inverted label setting.  Converesly, the method $\mathrm{JOINT} - \mathrm{PSEUDO}$ experiences more negative transfer between $M$ and $N$  than ADDA or our proposal, which we attribute to the instance-based nature of pseudo-label assignment (so is far more prone to noise between examples). Conversely,  the effects of internal covariate shift are apparent in $\bm{C}$ for   $\mathrm{REC}-\mathrm{MMD}(\mathbb{Q}_S \rightarrow \mathbb{Q}_T)$; in particular, the method only achieves 18.9\% on letter $S$.

\section{Conclusion} \label{sec:conclusion}

We extend adversarial discriminative domain adaptation for image classification by explicitly accounting for task knowledge in the discriminator during adversarial training and leveraging on the fixed distribution over source encoder posteriors.  In our proposal, we derive reconstruction and MMD loss formulations for adversarial training by    considering the discriminator as a denoising autoencoder with a reconstruction loss function and minimizing the maximum mean discrepancy between the discriminator posterior and source encoder posterior distribution, in order to train the encoder. We compare and analyze in detail how our method improves over conventional semi-supervised GAN loss formulations.  We also  introduce a simple regularization technique for reducing overfitting to the source domain for contraction mappings, which we intend on making adaptive to the domain adaptation scenario in future work. Our framework is shown to compete or outperform the state-of-the-art in unsupervised transfer learning on standard datasets, while remaining simple and intuitive to use. Finally, we show that our proposal minimizes the domain shift between emulated and real neuromorphic spike events on sign language recognition, improving substantially over source training only.

\appendices
\section{Analysis of Kernel Choice for $\mathcal{L}_T^\mathrm{MMD}(\mathbb{P}_S \rightarrow \mathbb{Q}_T)$} \label{App A}
We validate our choice of  RBF kernel function by considering other generalized variants.
The generalized RBF kernel can be expressed as follows: 
\begin{equation}
k_G =\exp\{- \rho D^2(\boldsymbol{x},\boldsymbol{y})\} 
\end{equation}

The constant $\rho$ is set to $\frac{1}{2\sigma_r}$ and $D^2$ is any  (conditionally) positive definite and symmetric distance.  In the paper, we set $D^2 = \left\Vert \boldsymbol{x}-\boldsymbol{y} \right\Vert^2_2$, the squared Euclidean ($L_2$) distance, which gives the standard RBF formulation. Given that our kernel function operates on  discrete probability distributions (i.e., the $K+1$-dimensional source encoder and target discriminator posteriors), this motivates testing generalized RBF kernels with probabilistic analogs for Euclidean distance. Namely, we consider the generalized RBF kernel with chi-squared distance ($D^2 = \sum_i\frac{(x_i-y_i)^2}{(x_i+y_i+ \epsilon)}$) and squared  Hellinger's distance ($D^2 = \sum_i(\sqrt{{x}_i+\epsilon}-\sqrt{{y}_i+\epsilon})^2$) - both generalized variants satisfy Mercer's condition and are typically used in conjunction with SVMs as a  non-linear mapping.   We set $\epsilon =10^{-8}$, as a fixed term to avoid division by 0 or undefined gradients at $\sqrt{0}$ for chi-squared and squared Hellinger's distance respectively. Finally, for completeness, we also consider the L1 distance, with $D^2=|\boldsymbol{x}-\boldsymbol{y}|$, as utilized in the Laplacian kernel.  

\begin{table} \centering \caption{Accuracy on SVHN $\rightarrow$ MNIST and MNIST $\rightarrow$ MNISTM tasks with varying $D^2$ }\begin{footnotesize} \resizebox{0.5\textwidth}{!} {\begin{tabular}{cccccccccccc} \toprule $D^2$ & MNIST $\rightarrow$ MNISTM & SVHN $\rightarrow$ MNIST \\

  \midrule
Chi-squared   & 0.918 & 0.898   \\

Sq. Hellinger & 0.890 & 0.909 \\

Absolute (L1) & 0.931 & 0.894 \\

Sq. Euclidean (L2) & \textbf{0.955} & \textbf{0.918} \\

\bottomrule 
 \end{tabular}} \end{footnotesize}  \label{tab:d_squared} \end{table} 

We present the results with varying kernel function in Table on SVHN $\rightarrow$ MNIST and MNIST $\rightarrow$ MNIST-M.   We use a summation over five kernels as in the paper, with  $\sigma_r = 10^{-r}, r \in \{0,\dots,4\},$ in order to constrain the computational cost. All other parameters  are fixed and we disable target regularization. As is evident from the Table \ref{tab:d_squared}, all kernels perform comparably but the standard RBF kernel with squared Euclidean distance achieves the best performance on both datasets.

\vspace{-8pt}

\bibliographystyle{IEEEtran}
\bibliography{egbib}

\vspace{-40pt}

\end{document}